\documentclass[preprint,5p,twocolumn]{elsarticle}

\biboptions{sort&compress}
\makeatletter
\def\ps@pprintTitle{%
	\let\@oddhead\@empty
	\let\@evenhead\@empty
	\let\@oddfoot\@empty
	\let\@evenfoot\@oddfoot
}
\makeatother
\usepackage[
type={CC},
modifier={by-nc-nd},
version={4.0},
]{doclicense}
\usepackage{overpic}
\usepackage[hyphens]{url}
\usepackage[hidelinks]{hyperref}
\usepackage{lineno}
\usepackage{optidef}
\usepackage{xcolor}
\usepackage{amsmath}
\usepackage{amssymb}
\usepackage{algpseudocode}
\usepackage{algorithm}
\usepackage{multirow}
\usepackage{filecontents}
\usepackage{pgfplots, pgfplotstable}
\usepgfplotslibrary{statistics}

\usepackage{tikz}
\usepackage{pgfplots}
\usetikzlibrary{calc}
\usepackage{bm}
\pgfplotsset{width=7cm,compat=1.14}
\usepgfplotslibrary{statistics}
\usetikzlibrary{positioning}
\usepackage{caption}
\usepackage{subcaption}
\usepackage[utf8]{inputenc}
\usetikzlibrary{arrows}
\usepackage{siunitx}
\usepgfplotslibrary{units}
\pgfplotsset{unit code/.code={\si{#1}}}
\usepackage{titlesec}
\usepackage{diagbox}
\usepackage{multirow}

\usepackage{etoolbox}
\makeatletter
\patchcmd\@combinedblfloats{\box\@outputbox}{\unvbox\@outputbox}{}{%
	\errmessage{\noexpand\@combinedblfloats could not be patched}%
}%
\makeatother

\usepackage{colortbl}
\usepackage{xcolor}
\definecolor{green}{RGB}{173,255,47}%
\definecolor{red}{RGB}{255, 0, 0}%
\definecolor{orange}{RGB}{229, 83, 0}%
\definecolor{redPlots}{RGB}{204,37,41}%
\definecolor{greenPlots}{RGB}{62,150,81}%
\definecolor{bluePlots}{rgb}{0.00000,0.44700,0.74100}%
\definecolor{turqoise}{rgb}{0.00000,0.64700,0.64100}%

\newcommand{\mr}[1]{\mathrm{#1}}

\newcommand{\R}{\mathbb{R}}
\newcommand{\X}{\mathbf{X}}

\newcommand{\vc}[1]{\mathbf{#1}}

\newcommand{\textred}[1]{{#1}}

\newcommand{\nin}{{n}}
\newcommand{\nout}{{m}}
\newcommand{\nhidn}{n}

\newcommand{\nts}{N}
\newcommand{\nruns}{T}

\newcommand{\hid}{z}

\newcommand{\ixhid}{k}
\newcommand{\ixts}{k}

\newcommand{\ixneuron}{i}

\titleformat{\paragraph}
{\normalsize\itshape}{\theparagraph}{1em}{}
\titlespacing*{\paragraph}
{0pt}{3.25ex plus 1ex minus .2ex}{1.5ex plus .2ex}

\newlength{\separ}

\newlength{\figW}
\newlength{\figH}

\usepackage[acronym,nopostdot,nonumberlist,nogroupskip,style=alttree]{glossaries}
\glssetwidest{ECMWF~}
\newacronym{ECMWF}{ECMWF}{European Center for Medium-Range Weather Forecasts}
\newacronym{DL}{DL}{Deep Learning}
\newacronym{DNN}{DNN}{Deep Neural Network}
\newacronym{NWP}{NWP}{Numerical Weather Prediction}
\newacronym{GHI}{GHI}{Global Horizontal Irradiance}
\newacronym{TPE}{TPE}{Tree-Structured Parzen Estimator}
\newacronym{rRMSE}{rRMSE}{Relative Root Mean Square Error}
\newacronym{SICSS}{SICSS}{Surface Insolation under Clear and Cloudy Skies}
\newacronym{GBT}{GBT}{Gradient Boosting Trees}
\newacronym{ARX}{ARX}{Autoregressive with Exogenous Inputs}
\newacronym{KNMI}{KNMI}{Royal Netherlands Meteorological Institute}

\makeglossaries
\begin{document}

\begin{frontmatter}

\title{Short-term forecasting of solar irradiance without local telemetry: a generalized model using satellite data}

\author[delft,vito,cor1]{Jesus Lago}
\ead{j.lagogarcia@tudelft.nl}
\author[3e]{Karel De Brabandere}
\author[vito]{Fjo De Ridder}
\author[delft]{Bart De Schutter}
\address[delft]{Delft Center for Systems and Control, Delft University of Technology,\\ 
	Mekelweg 2, Delft, The Netherlands}
\address[vito]{Algorithms, Modeling, and Optimization, VITO, Energyville, \\
	ThorPark, Genk, Belgium}
\address[3e]{3E, Brussels, Belgium}
\cortext[cor1]{Corresponding author}
\tnotetext[t1]{This is the postprint of the article: \textit{Short-term forecasting of solar irradiance without local telemetry: a generalized model using satellite data, Solar Energy 173 (2018), 566-577}. \url{https://doi.org/10.1016/j.solener.2018.07.050}.}
\begin{abstract}
Due to the increasing integration of solar power into the electrical grid, forecasting short-term solar irradiance has become key for many applications, e.g.~operational planning, power purchases, reserve activation, etc. In this context, as solar generators are geographically dispersed and ground measurements are not always easy to obtain, it is very important to have general models that can predict solar irradiance without the need of local data. In this paper, a model that can perform short-term forecasting of solar irradiance in any general location without the need of ground measurements is proposed. To do so, the model considers satellite-based measurements and weather-based forecasts, and employs a deep neural network structure that is able to generalize across locations; \textred{particularly, the network is trained only using a small subset of sites where ground data is available, and the model is able to generalize to a much larger number of locations where ground data does not exist.} As a case study, 25 locations in The Netherlands are considered and the proposed model is compared against four local models that are individually trained for each location using ground measurements. Despite the general nature of the model, it is shown show that the proposed model is equal or better than the local models: when comparing the average performance across all the locations and prediction horizons, the proposed model obtains a 31.31\% rRMSE (relative root mean square error) while the best local model achieves a 32.01\% rRMSE.
\end{abstract}


\end{frontmatter}

\printglossaries

\section{Introduction}
\glsaddall
With the increasing integration of renewable sources into the electrical grid, accurate forecasting of renewable source generation has become one of the most important challenges across several applications. Among them, balancing the electrical grid via activation of reserves is arguably one of the most critical ones to ensure a stable system. In particular, due to their intermittent and unpredictable nature, the more renewables are integrated, the more complex the grid management becomes \cite{Lara-Fanego2012,Voyant2017}.

In this context, as solar energy is one of the most unpredictable renewable sources, the increasing use of solar power in recent years has led to an increasing interest in forecasting irradiance over short time horizons. In particular, in addition to activation of reserves to manage the grid stability, short-term forecasts of solar irradiance are paramount for operational planning, switching sources, programming backup, short-term power trading, peak load matching, scheduling of power systems, congestion management, and cost reduction \cite{Hammer1999,Reikard2009,Voyant2017}.

\subsection{Solar irradiance forecasting}
The forecasting of solar irradiance can be typically divided between methods for \textit{global horizontal irradiance (GHI)} and methods for
\textit{direct normal irradiance (DNI)} \cite{Law2014}, with the latter being a component of the GHI (together with the diffuse solar irradiance). As in this work GHI is forecasted, \cite{Law2014} should be used for a complete review on methods for DNI. For the case of GHI, forecasting techniques are further categorized into two subfields according to the input data and the forecast horizon \cite{Diagne2013,Voyant2017}:

\begin{enumerate}
\item Time series models based on satellite images,
measurements on the ground level, or sky images. These methods are usually suitable for short-term forecasts up to 4-6 h. Within this field, the literature can be further divided into three groups.
\begin{enumerate}
	\item Classical statistical models like ARMA models \cite{Ahmad2015}, ARIMA models \cite{Reikard2009}, the CARDS model \cite{Huang2013}, or the Lasso model \cite{Yang2015}.
	\item Artificial intelligence models such as neural networks models \cite{Mellit2010,Lauret2015}, support vector machines \cite{Lauret2015}, decision trees-based models \cite{McCandless2015}, or Gaussian models \cite{Lauret2015}.
	\item Cloud-moving vector models that use satellite images \cite{Lorenz2012}.
\end{enumerate}
\item \textit{Numerical weather prediction (NWP)} models that simulate weather conditions. These methods are suitable for longer forecast horizons, 4-6 hours onward, time scales where they outperform the statistical models \cite{Perez2010}. As the goal of this work are short-term forecasts, \cite{Diagne2013} should be used for more complete review of NWP methods.
\end{enumerate}

\noindent While the division in accuracy between NWP and time series models is given by the predictive horizon, establishing comparisons between time series models is more complex. In particular, while some authors have reported the superiority of statistical models over artificial intelligence methods \cite{Reikard2009}, others have obtained opposite results \cite{Sfetsos2000}.

 \textred{The input features typically used in the literature to predict solar irradiance vary widely, e.g.~past irradiance values, satellite data, weather information, etc. In many cases, the inputs considered depend on the type of model used, e.g.~cloud moving vector models require satellite images. While a detailed review on the different methods and input features is outside the scope of this paper, \cite{Diagne2013} is a good source for a more thorough analysis.}

\subsection{Motivation}
To the best of our knowledge, due to the time series nature of the solar irradiance, the statistical and artificial intelligence methods proposed so far have considered past ground measurements of the solar irradiance as input regressors \cite{Diagne2013}. While this choice of inputs might be the most sensible selection to build time series models, it poses an important problem: local data is required at every site where a forecast is needed.

In particular, if the geographical dispersion of solar generators is considered, it becomes clear that forecasting solar irradiance is a problem that has to be resolved across multiple locations. If ground measurements of all these sites are required, the cost of forecasting irradiance can become very expensive. In addition to the cost, a second associated problem is the fact that obtaining local data is not always easy. 

As a result, in order to obtain scalable solutions for solar irradiance forecasting, it is important to develop global models that can forecast without the need of local data. In this context, while current cloud-moving vectors might accomplish that, they are not always easy to deploy as they are complex forecasting techniques that involve several steps \cite{Diagne2013}.

\subsection{Contributions and Organization of the Paper}
In this paper, a novel forecasting technique is proposed that addresses the mentioned problem by providing a prediction model that, while being accurate and easy to deploy, forecasts solar irradiance without the need of local data. The prediction model is based on a \textit{deep neural network (DNN)} that, using SEVIRI\footnote{The SEVIRI (Spinning Enhanced Visible and InfraRed Imager) is a measurement instrument of the METEOSAT satellite.} satellite images and NWP forecasts, is as accurate as local time series models that consider ground measurements. Although the model uses satellite images just as cloud-moving vector models do, it is easier to deploy as it requires less complex computations. \textred{In addition, while obtaining satellite data might not be always easier or cheaper than installing local ground sensors, there are several locations where satellite data are available and the proposed model avoids going to the ground to install local measurements. An example of this is The Netherlands, where satellite data is provided by the national meteorological institute.}

\textred{It is important to note that, to the best of our knowledge, the proposed method is the first of its class that tries to remove the dependence of local telemetry even for training. Particularly, while other methods from the literature successfully remove the local data dependence during forecasting, e.g.~\cite{Larson2018}, they still require local telemetry at all sites of interest during training. While using local data in a small subsets of sites during training, the proposed model successfully predicts the irradiance in a much larger subset of locations without needing local telemetry from these sites at any stage of the estimation or the forecasting.
}

As a case study, 30 location in The Netherlands are considered and the model is estimated using 5 of these locations. Then, for the remaining 25 locations, the performance of the proposed estimated model is compared against individual time series models specifically trained for each site using ground data.

The remaining of the paper is organized as follows: Section \ref{sec:theory} introduces the preliminary concepts considered in this work. Next, Section \ref{sec:modelframe} presents the proposed general model for forecasting solar irradiance. Then, Section \ref{sec:casestudy} introduces the case study and discusses the performance of the proposed model when compared with local models. Finally, Section \ref{sec:conclusion} summarizes the main results and concludes the paper.

\section{Preliminaries}
\label{sec:theory}
In this section the concepts and algorithms that are used and/or modified in the paper are introduced.

\subsection{Deep Learning and DNNs}
\label{sec:deeplearning}
In the last decade, the field of neural networks has experienced several innovations that have lead to what is known as \textit{deep learning (DL)} \cite{Goodfellow2016}. In particular, one of the traditional issues of neural networks had always been the large computational cost of training large models. However, that changed completely when \cite{Hinton2006} showed that a deep belief network could be trained efficiently using an algorithm called greedy layer-wise pretraining. As related developments followed, researchers started to be able to efficiently train complex neural networks whose depth was not just limited to a single hidden layer (as in the traditional multilayer perceptron). As these new structures systemically showed better results and generalization capabilities, the field was renamed as deep learning to stress the importance of the depth in the achieved improvements \cite[Section~1.2.1]{Goodfellow2016}.

While this success of DL models initiated in computer science applications, e.g.~image recognition \cite{Krizhevsky2012}, speech recognition \cite{Hinton2012}, or  machine translation \cite{Bahdanau2014}, the benefits of DL have also spread in the last years to several energy-related applications \cite{Wang2016,Feng2017,Suryanarayana2018,Coelho2017,Fan2017,Lago2018,Lago2018a}. Among these areas, wind power forecasting \cite{Wang2016,Feng2017} and electricity price forecasting \cite{Lago2018,Lago2018a} are arguably the fields that have benefited the most

While there are different DL architectures, e.g.~convolutional networks or recurrent networks, in this paper a DNN is considered, i.e.~a multilayer perceptron with more than a single hidden layer, in order to build the solar forecasting model. {The reason for this selection is twofold: (1) DNNs are less computationally intensive than the other DL architectures \cite{Goodfellow2016}; (2) DNNs have empirically outperformed the other DL architectures in a similar energy-based forecasts \cite{Lago2018a}, i.e.~the forecast of day-ahead electricity prices.
}
\subsubsection{Representation}
\begin{figure}[htb]
	\begin{center}
		\def\Nin{2}
\def\Nhid{2}
\def\Nhidd{2}
\def\Nout{2}
\setlength{\separ}{2cm}
\begin{tikzpicture}[shorten >=1pt,->,draw=black!50]

\tikzstyle{every pin edge}=[<-,shorten <=1pt]
\tikzstyle{neuron}=[circle,fill=black!25,minimum size=17pt,inner sep=0pt]
\tikzstyle{input neuron}=[neuron, fill=redPlots!60];
\tikzstyle{output neuron}=[neuron, fill=greenPlots!80];
\tikzstyle{hidden neuron}=[neuron, fill=orange!70];
\tikzstyle{hidden neuron 2}=[neuron, fill=bluePlots!70];
\tikzstyle{annot} = [text width=4em, text centered]

\foreach \name / \y in {1,...,\Nin}
\node[input neuron] (I-\name) at (0,-\y) {$x_{\y}$};
\node at (0,-\Nin-1) {$\vdots$};
\node[input neuron] (I-\Nin+1) at (0,-\Nin-2) {$x_{\nin}$};

\foreach \name / \y in {1,...,\Nhid}
\path[yshift=0cm]
node[hidden neuron] (H-\name) at (\separ,-\y cm) {$\hid_{1\y}$};
\node at (\separ,-\Nhid-1) {$\vdots$};
\node[hidden neuron] (H-\Nhid+1) at (\separ,-\Nhid-2) {$\hid_{1\nhidn_1}$};

\foreach \name / \y in {1,...,\Nhidd}
\path[yshift=0cm]
node[hidden neuron 2] (H2-\name) at (2\separ,-\y cm) {$\hid_{2\y}$};
\node at (2\separ,-\Nhidd-1) {$\vdots$};
\node[hidden neuron 2] (H2-\Nhidd+1) at (2\separ,-\Nhidd-2) {$\hid_{2 \nhidn_2}$};

\foreach \name / \y in {1,...,\Nout}
\path[yshift=0cm]
node[output neuron] (O-\name) at (3\separ,-\y cm) {$y_{\y}$};
\node at (3\separ,-\Nout-1) {$\vdots$};
\node[output neuron] (O-\Nout+1) at (3\separ,-\Nout-2) {$y_{{\nout}}$};



\foreach \source in {1,...,\Nin}
\foreach \dest in {1,...,\Nhid}
\path (I-\source) edge (H-\dest);

\foreach \dest in {1,...,\Nhid}
\path (I-\Nin+1) edge (H-\dest);
\foreach \source in {1,...,\Nin}
\path (I-\source) edge (H-\Nhid+1);
\path (I-\Nin+1) edge (H-\Nhid+1);

\foreach \source in {1,...,\Nhid}
\foreach \dest in {1,...,\Nhidd}
\path (H-\source) edge (H2-\dest);

\foreach \dest in {1,...,\Nhidd}
\path (H-\Nhid+1) edge (H2-\dest);
\foreach \source in {1,...,\Nhid}
\path (H-\source) edge (H2-\Nhidd+1);
\path (H-\Nhid+1) edge (H2-\Nhidd+1);

\foreach \source in {1,...,\Nhidd}
\foreach \dest in {1,...,\Nout}
\path (H2-\source) edge (O-\dest);

\foreach \dest in {1,...,\Nout}
\path (H2-\Nhidd+1) edge (O-\dest);
\foreach \source in {1,...,\Nhid}
\path (H2-\source) edge (O-\Nout+1);
\path (H2-\Nhidd+1) edge (O-\Nout+1);

\node[annot,above of=H-1, node distance=0.75cm] (hl) {Hidden layer};
\node[annot,above of=H2-1, node distance=0.75cm] (hl) {Hidden layer};
\node[annot,above of=I-1, node distance=0.75cm] {Input layer};
\node[annot,above of=O-1, node distance=0.75cm] {Output layer};

\node at (3\separ,-\Nout-1) {$\vdots$};
\end{tikzpicture}
		\caption{Example of a DNN.}
		\label{fig:hiddnetexa}
	\end{center}
\end{figure}
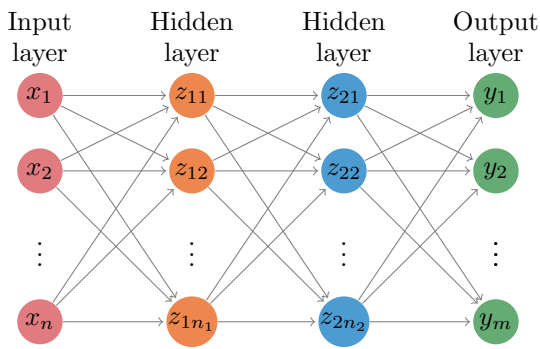

Defining by $\mathbf{X}=[x_1,\ldots,x_\nin]^\top\in\R^\nin$ the input of the network, by $\mathbf{Y}=[y_{{1}},y_{{2}},\ldots,y_{{\nout}}]^\top\in\R^{\nout}$ the output of the network, by $\nhidn_\ixhid$ the number of neurons of the $\ixhid^\mr{th}$ hidden layer, and by $\mathbf{\hid}_\ixhid=[\hid_{\ixhid1},\ldots,\hid_{\ixhid \nhidn_\ixhid}]^\top$ the state vector in the ${\ixhid}^\mr{th}$ hidden layer, a general DNN with two hidden layers can be represented as in Figure \ref{fig:hiddnetexa}.

In this representation, the parameters of the model are represented by the set of parameters $\vc{W}$ that establish the mapping connections between the different neurons of the network \cite{Goodfellow2016}.

\subsubsection{Training}

The process of estimating the model weights $\vc{W}$ is usually called training. In particular, given a training set $\mathcal{S_T}=\bigl\{(\X_\ixts,\vc{Y}_\ixts)\bigr\}_{\ixts=1}^\nts$ with $N$ data points, the network training is done by solving a general optimization problem with the following structure:
\begin{mini}
	{\vc{W}}{\sum_{\ixts=1}^{\nts}g_\ixts\Bigl(\vc{Y}_\ixts, F(\vc{X}_\ixts,\vc{W})\Bigr),}
	{\label{eq:prob1}}{}
\end{mini}

\noindent where $F:\R^\nin\rightarrow\R^{\nout}$ is the neural network map, and $g_\ixts$ is the problem-specific cost function, e.g.~the Euclidean norm or the average cross-entropy. Traditional methods to solve \eqref{eq:prob1} include the {gradient descent} or the {Levenberg–Marquardt} algorithm \cite{Weron2014}. However, while these methods work well for small sized-networks, they display computational and scalability issues for DNNs. In particular, for DNNs better alternatives are the {stochastic gradient descent algorithm} and all its variants \cite{Ruder2016}.

{It is important to note that \eqref{eq:prob1} is an approximation of the real problem one wish to solve. Particularly, in an ideal situation, the cost function w.r.t.~to the underlying data distribution would be minimized; however, as the distribution is unknown, the problem has to be approximated by minimizing the cost function over the finite training set. This is especially relevant for neural networks, where a model could be overfitted and have a good performance in the training set, but perform badly in the test set, i.e.~a set with a different data distribution. To avoid this situation, the network is usually trained in combination with regularization techniques, e.g.~early stopping, and using out-of-sample data to evaluate the performance \cite{Goodfellow2016}.}

\subsubsection{Network Hyperparameters}
In addition to the weights, the network has several parameters that need to be selected before the training process. Typical parameters include the number of neurons of the hidden layers, the number of hidden layers, or the learning rate of the stochastic gradient descent method. To distinguish them from the main parameters, i.e.~the network weights, they are referred to as the network hyperparameters.

\subsection{Hyperparameter Optimization and Feature Selection}
\label{sec:hyper}
In this paper, to perform the hyperparameter selection, a Bayesian optimization algorithm that has been widely used for hyperparameter selection is considered: the \textit{tree-structured Parzen estimator (TPE)} \cite{Bergstra2011}, an optimization algorithm within the family of {sequential model-based optimization} methods \cite{Hutter2011}. The basic principle of a sequential model-based optimization algorithm is to optimize a black-box function, e.g.~the performance of a neural network as a function of the hyperparameters, by iteratively estimating an approximation of the function and exploring the function space using the local minimum of the approximation. At any given iteration $i$, the algorithm evaluates the black-box function at a new point $\bm{\theta}_i$. Next, it estimates an approximation $\mathcal{M}_i$ of the black-box function by fitting the previously sampled points to the obtained function evaluations. Then, it selects the next sample point $\bm{\theta}_{i+1}$ by numerically optimizing $\mathcal{M}_i$ and starts the next iteration. Finally, after a maximum number of iterations $\nruns$ have been performed, the algorithm selects the best configuration. 	
Algorithm \ref{al:smbo} represents an example of a sequential model-based optimization algorithm for hyperparameter selection.
\begin{algorithm}
	\caption{Hyperparameter Optimization}
	\label{al:smbo}
	\begin{algorithmic}[1]
		\Procedure{SMBO}{$\nruns,\bm{\theta}_1$}
		\State $\mathcal{H} \gets \emptyset $
		\For{$i=1,\ldots,\nruns$}
		\State $ p_i \gets$ TrainNetwork($\bm{\theta}_i$) 
		\State $\mathcal{H} \gets \mathcal{H} \cup \bigl\{(p_i,\bm{\theta}_i)\bigr\}$
		\If{$i < T$}
		\State $\mathcal{M}_i(\bm{\theta}) \gets \mr{EstimateModel}(\mathcal{H})$
		\State $ \bm{\theta}_{i+1} \gets \mr{argmax}_{\bm{\theta}} ~\mathcal{M}_i(\bm{\theta})$
		\EndIf
		\EndFor
		\State $\bm{\theta}^* \gets \mr{BestHyperparameters}(\mathcal{H})$
		\State \Return $\bm{\theta}^*$
		\EndProcedure
	\end{algorithmic}
\end{algorithm}

{In addition to optimizing the hyperparameters, the TPE algorithm is also employed for optimizing the selection of input features. In particular, the feature selection method proposed in \cite{Lago2018} is considered, which selects the input features by first defining the input features as model hyperparameters and then using the TPE algorithm to optimally choose among them. More specifically, the method considers that each possible input feature can be either modeled as a binary hyperparameter representing its inclusion/exclusion or as an integer hyperparameter representing how many historical values of the specific input are used. In solar forecasting, an example of the former could be whether to consider the hour of the day as an input feature and an example of the latter could be the optimal number of past irradiance values.}

\subsection{Performance Metrics}
\label{sec:metrics}
\textred{In order to evaluate the accuracy of the proposed model, a performance metric is needed. In this paper, following the standards of the literature of solar irradiance forecasting, three different metrics are considered: the \textit{relative root mean square error (rRMSE)}, the the \textit{mean bias error (MBE)}, and the forecasting skill $s$ as defined by \cite{Marquez2012}.

One of the most commonly used metrics for evaluating solar irradiance forecasting is the RMSE or rRMSE, which provide an assessment of the average spread of the forecasting errors. In particular, given a vector $\vc{Y}=[y_1,\ldots,y_\nts]^\top$ of real outputs and a vector $\vc{\hat{Y}}=[\hat{y}_1,\ldots,\hat{y}_\nts]^\top$ of predicted outputs, the rRMSE metric can be computed as:
\begin{equation}
\mathrm{rRMSE} =  \frac{\sqrt{\frac{1}{N}\sum_{\ixts=1}^{\nts}(y_\ixts-\hat{y_\ixts})^2}}{\frac{1}{N}\sum_{\ixts=1}^{\nts} y_\ixts}\cdot 100\,\,\%.
\end{equation}

A second metric that is widely used is the MBE, a measure of the overall bias of the model. Using the same definitions as before, the MBE metric can be computed as:

\begin{equation}
\frac{1}{N}\sum_{\ixts=1}^{\nts}y_\ixts-\hat{y_\ixts}.
\end{equation}

While both metrics can properly assess and compare models using the same dataset, they are hard to interpret when it comes to make comparisons across multiple locations, climate, and time of the year \cite{Marquez2012}. A metric that tries to solve this issue is the forecasting skill $s$; particularly, $S$ defines first a metric $V$ that accounts for the \textit{variability} of the solar irradiance, i.e.~accounts for the specific variability due to location, climate, and time. Next, it defines a second metric $U$ that accounts for the \textit{uncertainty}, i.e.~errors, of the forecasting model. Finally, the forecasting skill $S$ is defined as:

\begin{equation}
	s = 1-\frac{U}{V}.
\end{equation}

\noindent For the details on computing $U$ and $V$ as well as a detailed explanation on $s$, the reader is referred to \cite{Marquez2012}. The important aspect to consider for this study is that $s$ is a normalized metric w.r.t.~to a simple persistence model (see Section \ref{sec:pers}) that permits the comparison of models across different conditions. A normal forecaster should be characterized by $s\in[0,1]$ with higher values indicating better
forecasting; particularly, $s=1$ indicates that the solar irradiance is perfectly forecasted, and $s=0$ that the model is not better than a simple persistence model (by definition of $U$ and $V$ a persistence model will always have $s=0$). Negative values would then imply the forecaster is worse than the simple persistence model.
}

\section{Prediction Model}
\label{sec:modelframe}
In this section, the proposed prediction model for solar irradiance forecasting is presented. 
\subsection{Model Structure}
\label{sec:structure}
A key element to build a prediction model that can be used without the need of ground data is to employ a model whose structure is flexible enough to generalize across multiple geographical locations. As DNNs are powerful models that can generalize across tasks \cite{Goodfellow2016,Lago2018}, they are selected as the base model for the proposed forecaster. This concept of generalization is further explained in Section \ref{sec:modelGen}.

While the model is a DNN as the one illustrated in Figure \ref{fig:hiddnetexa}, the number of layers, the size of the output, and the type of inputs are specifically selected according to the application. In particular, considering that 6 hours is the limit predictive horizon before NWP forecast outperform time series models \cite{Diagne2013}, the model consists of $6$ output neurons representing the forecasted hourly irradiance over the next 6 hours; this horizon is the standard choice for short-term irradiance forecasting \cite{Diagne2013}.

In terms of hidden layers, the model is not subject to any specific depth; instead, depending on the case study, i.e.~the geographical area where the forecasts are made, the number of hidden layers are optimized using hyperparameter optimization as explained in Sections \ref{sec:hyper}. For the case study in this paper, i.e.~forecasting irradiance in the Netherlands, the optimal network depth is 2 hidden layers. To select the number of neurons per layer, the same methodology applies, i.e.~they need to be optimized for each geographical location.

\subsection{Model Inputs}
\label{sec:inputs}
As indicated in the introduction, the aim of the model is to forecast solar irradiance without the need of ground data. As a result, to perform the selection of model inputs, it is paramount to consider the subset of inputs that, while correlating with solar irradiance, are general enough so that they can be easily obtained for any given location. Given that restriction, the proposed model considers three types of inputs: NWP forecasts of the solar irradiance, the clear-sky irradiance, and satellite images representing maps of past solar irradiance.

\subsubsection{Numerical weather prediction forecast}
The first type of input are NWP forecasts of the solar irradiance obtained from the \textit{European center for medium-range weather forecasts (ECMWF)}. As indicated in the introduction, NWP forecasts of the solar irradiance are less accurate than time series models for short-term horizons. However, as they strongly correlate with the real irradiance, they are very useful regressors to build time series models.

For the proposed model, the input data consists of the 6 forecasted values for the next 6 hours given by the latest available ECMWF forecast (typically available every day around 08:00-09:00 CET).

\subsubsection{Clear-sky irradiance}
As second input, the model considers the clear-sky irradiance ${\mathcal{I}}_\mathrm{c}$, i.e.~the GHI under clear-sky conditions, at every hour over the next 6 hours. The clear-sky irradiance is a deterministic input that is obtained using the clear-sky model defined in \cite{Ineichen2002}, which computes ${\mathcal{I}}_\mathrm{c}$ using the location and time of interest.

\subsubsection{Satellite images}
\label{sec:sat}
The third input are satellite data representing the past irradiance values of a geographical area. In particular, the input data consists of images from the SEVIRI instrument of the METEOSAT satellite that are transformed to irradiance values using two different methods:
\begin{enumerate}
	\item For data corresponding to solar elevation angles above 12$^\circ$, the SEVIRI-based images are mapped to irradiance values using the \textit{Surface insolation under clear and cloudy skies (SICSS)}  algorithm \cite{Greuell2013}. 
	\item For data corresponding to solar elevation angles below 12$^\circ$, i.e.~very early in the morning and late in the evening, the irradiance values are extracted by considering the interpolation method described in \cite{Deneke2008} applied to the clear sky index.
\end{enumerate}
This distinction depending on the solar elevation angle is required because: (1) the SICSS method considers cloud properties; (2) at low solar elevation angles the uncertainty in the cloud properties increases strongly \cite{Deneke2008}.

Once the satellite images are mapped to irradiance values, the input data simply consists of the past irradiance values in the individual pixel where the forecasting site is located. Then, to select which past irradiance values, i.e.~which past images, are relevant for building the general model, the feature selection method defined in Section \ref{sec:hyper} is employed.

As a final remark, it is important to note that these irradiance values have a resolution that is limited by the resolution of the satellite images, which in the case of the SEVIRI instrument are pixels of $3\times 3$\,km. 
As a result, to represent the solar irradiance in a specific location, the accuracy of satellite-based measurements cannot be better than that of ground measurements. 


\textred{\subsubsection{Input selection}
	\label{sec:inpselection}
The three input features that the proposed model considers were selected from a larger set of input features. In particular, in order to ensure that the proposed model included the most relevant input features, a feature selection process was performed. During this feature selection process, the three considered inputs, i.e.~the NWP forecasts, the clear-sky irradiance, and the satellite images were selected as the most important features. However, in addition to these three, four other features were also considered:

\begin{itemize}
	\item Historical values of the temperature.
	\item Historical values of the humidity.
	\item Forecast of the temperature.
	\item Forecast of the humidity.
\end{itemize}

To perform the feature selection between these 7 input features, the feature selection method described in \cite{Lago2018} was employed; i.e.~the 7 input features were modeled as binary hyperparameters and the selection was performed together with the hyperparameter optimization described in Section \ref{sec:modehyper}. This optimization resulted in the 3 selected inputs.
}

\subsection{Hyperparameter Optimization and Feature Selection}
\label{sec:modehyper}
As briefly introduced in Section \ref{sec:structure}, the proposed model needs to be tuned for the specific geographical area where it is applied. In order to tune the model structure, the following four DNN hyperparameter are optimized:
\begin{enumerate}
	\item \textbf{Number of hidden layers}: the neural network depth is a parameter that needs to be tuned in order to obtain a model that can correctly generalize across multiple geographical locations.
	\item \textbf{Number of neurons per layer}: besides the number of hidden layers, the size of each layer also plays an important role in the generalization capabilities of the DNN.
	\item \textbf{General learning rate}: the initial learning rate used in the stochastic gradient descent method. {In particular, while the stochastic gradient descent method automatically adapts the learning rate at every iteration of the optimization process, the learning rate at the first iteration has to be selected.}
	\item \textbf{Dropout}: Dropout \cite{Srivastava2014} is included as a possible regularization technique to reduce overfitting and to improve the training performance. To do so, at each iteration, dropout selects a fraction of the neurons and prevents them from training. This fraction of neurons is defined as a real hyperparameter between 0 and 1.
\end{enumerate}

\textred{As explained in Section \ref{sec:hyper} and \ref{sec:inpselection}, in combination with the hyperparameter optimization, the proposed model also performs a feature selection. In particular, the feature selection method selects the most relevant inputs among a subset of 7 features and it also selects which past historical irradiance values are required.}

\subsection{Model Parameters}
The parameters of the DNN are represented by the set of weights that establish the mapping connections between the several neurons of the network:

\begin{itemize}
	\item $\mathbf{W}_{\mr{i},\ixneuron}$: the vector of weights between the input $\X$ and the neuron $\ixneuron$ of the first hidden layer.
	\item $\mathbf{W}_{{k, \ixneuron}}$: the vector of weights between the $k^\mr{th}$ hidden layer and the neuron $\ixneuron$ of the $(k+1)^\mr{th}$ hidden layer.
	\item $\mathbf{W}_{\mr{o},\ixneuron}$: the vector of weights between the last hidden layer and the irradiance price vector $\hat{\mathcal{I}}$.
	\item $\mathbf{b}_\ixhid =[b_{\ixhid1},\ldots,b_{\ixhid{\nhidn_\ixhid}}]^\top$: the vector of bias weights in the ${\ixhid }^\mr{th}$ hidden layer, with $k=1,\,2$.
	\item $\mathbf{b}_\mr{o}=[b_{\mr{o},1}\ldots,b_{\mr{o},6}]^\top$: the vector of bias weights in the output layer.
\end{itemize}

\subsection{Model Equations}
Using the above definitions, the equations of the DNN assuming two hidden layers can be defined as:
\begin{subequations}
	\label{eq:deepnn}
	\begin{alignat}{3}
	\!\!\!\!\hid_{1\ixneuron} &= f_{1\ixneuron}\Bigl(\mathbf{W}_{\mr{i},\ixneuron}^\top \cdot \mathbf{X}+b_{1\ixneuron}\Bigr),\quad &&\mr{for~}\ixneuron=1,\ldots \nhidn_1,\\
	\hid_{2\ixneuron} &= f_{2\ixneuron}\Bigl(\mathbf{W}_{{2\ixneuron}}^\top \cdot  \mathbf{\hid}_1+b_{2\ixneuron}\Bigr),\quad&&\mr{for~}\ixneuron=1,\ldots \nhidn_2,\\
	\!\!\!\! \hat{\mathcal{I}}_{h+i} &= \mathbf{W}_{\mr{o},\ixneuron}^\top \cdot \mathbf{\hid}_2+b_{\mr{o},\ixneuron},\quad &&\mr{for~} \ixneuron=1,\ldots 6, \label{eq:genoutDNN}
	\end{alignat}
\end{subequations}

\noindent where $f_{k \ixneuron}$ represents the activation function of neuron $\ixneuron$ in the $k^\mr{th}$ hidden layer. In particular, for the proposed model, the \textit{rectified linear unit (ReLU)} \cite{Nair2010} is selected as the activation function of the two hidden layers. {This choice is made because this activation function has become a standard for hidden layers of DNNs \cite{Goodfellow2016}.} It is important to note that, as the irradiance is a real number, no activation function is used for the output layer. 



\subsection{Training
}
\label{sec:singledetails}
The DNN is trained  by minimizing the mean square error\footnote{Note that minimizing the mean square error is equivalent to minimizing the rRMSE metric used throughout the paper to evaluate and compare the model.}. In particular, given the training set $\mathcal{S_T}=\bigl\{(\X_\ixts,\hat{\mathbf{I}}_\ixts)\bigr\}_{\ixts=1}^\nts$, the optimization problem that is solved to train the neural network is:
\begin{mini}
	{\vc{W}}{\sum_{\ixts=1}^{\nts}\|\hat{\mathbf{{I}}}_\ixts - F(\vc{X}_\ixts,\vc{W})  \|_2^2,}
	{}{}
\end{mini}

\noindent where $F:\R^\nin\rightarrow\R^{6}$ is the neural network map and $\mathbf{W}$ is the set comprising all the n weights and bias weights of the network.

\subsubsection{Generalizing across geographical sites}
\label{sec:modelGen}
A key element for the model to forecast without the need of ground data is to be able to generalize across locations. To do so, the proposed model is trained across a small subset of sites so that the model learns to generalize across geographical sites. It is important to note that, while ground data is required for this small subset of locations, the model generalizes across all other geographical locations where ground data is not needed. In particular, as it is shown in the case study for The Netherlands, the number of locations where ground data is required is relatively small, e.g.~3-5 sites.

\subsubsection{Generalizing across predictive horizons}
Enforcing generalization is not only good for obtaining a model that does not require ground data, but in general, it is also beneficial to obtain a DNN that does not overfit and that obtains more accurate predictions \cite{Goodfellow2016}. In particular, as it has been empirically shown in several studies \cite{Lago2018,Lago2018a}, by forcing the network to solve multiple related task, e.g.~forecasting multiple sites, the network might learn to solve individual tasks better.

Therefore, to further strengthen the generalization capabilities of the network, the DNN is trained to forecast over the next 6 hours but starting at any hour of the day. As with the geographical site generalization, the goal is to build a DNN that, by performing several related tasks, it is able to learn more accurate predictions.

\subsubsection{Implementation details}
{The optimization problem is solved using multi-start optimization and Adam \cite{Kingma2014}, a version of stochastic gradient descent that computes adaptive learning rates for each model parameter. The use of adaptive learning rates is selected for a clear reason: as the learning rate is automatically computed, the time needed to tune the learning rate is smaller in comparison with other optimization methods. Together with Adam, the forecaster also considers early stopping \cite{Yao2007} to avoid overfitting.} 

\textred{\subsection{Issues}
	Note that the proposed model depends on another type of forecasts provided by NWP models. As a consequence, if the NWP models are performing bad, they might impact the final performance of the prediction model. For the proposed model, one of the most accurate and well-known NWP forecast models is considered: the ECMWF forecast \cite{ecmwf}. If other NWP models are employed instead, the performance of the model might vary w.r.t.~the results shown in this paper."
}

\subsection{Representation}

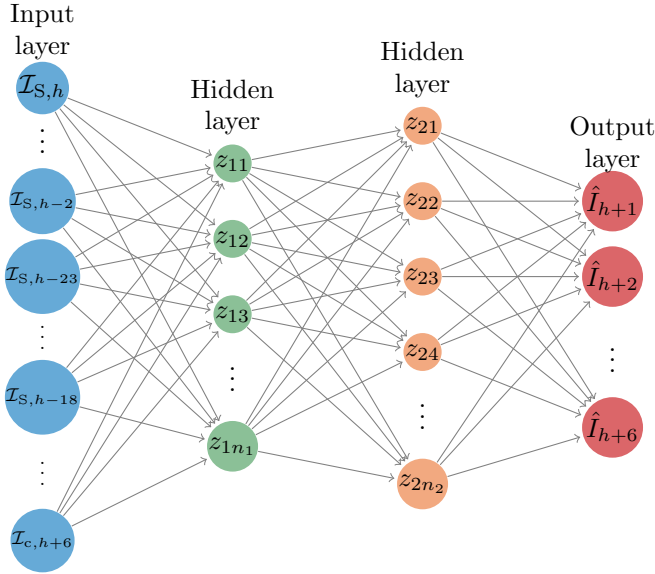
\begin{figure*}[htb]
	\begin{center}
		\def\Nin{2}
\def\Nhid{3}
\def\Nhidd{4}
\def\Nhiddd{5}
\def\Nhidddd{3}
\def\Nout{2}
\setlength{\separ}{2.5cm}
\begin{tikzpicture}[shorten >=1pt,->,draw=black!50]

\tikzstyle{every pin edge}=[<-,shorten <=1pt]
\tikzstyle{neuron}=[circle,fill=black!25,minimum size=12pt,inner sep=0pt]
\tikzstyle{input neuron}=[neuron, fill=bluePlots!60];
\tikzstyle{output neuron}=[neuron, fill=redPlots!70];
\tikzstyle{hidden neuron}=[neuron, fill=greenPlots!60];
\tikzstyle{hidden neuron 2}=[neuron, fill=orange!50];
\tikzstyle{hidden neuron 3}=[neuron, fill=greenPlots!60];
\tikzstyle{hidden neuron 4}=[neuron, fill=bluePlots!30];
\tikzstyle{annot} = [text width=4em, text centered]

\node[input neuron] (I-1) at (0,0.5) {$\mathcal{I}_{\mathrm{S},h}$};
\node at (0,-0.1) {$\vdots$};
\scriptsize
\node[input neuron] (I-12) at (0,-1) {$\mathcal{I}_{\mathrm{S},h-2}$};
\node[input neuron] (I-13) at (0,-2) {$\mathcal{I}_{\mathrm{S},h-23}$};
\node at (0,-2.7) {$\vdots$};
\node[input neuron] (I-2) at (0,-3.6) {$\mathcal{I}_{\mathrm{S},h-18}$};
\node at (0,-4.5) {$\vdots$};
\node[input neuron] (I-\Nin+1) at (0,-\Nin-3.5) {$\mathcal{I}_{\mathrm{c},h+6}$};
\normalsize
\foreach \name / \y in {1,...,\Nhid}
\path[yshift=0.5cm]
node[hidden neuron] (H-\name) at (\separ,-\y cm) {$\hid_{1\y}$};
\node at (\separ,-\Nhid-0.25) {$\vdots$};
\node[hidden neuron] (H-\Nhid+1) at (\separ,-\Nhid-1.25) {$\hid_{1\nhidn_1}$};

\foreach \name / \y in {1,...,\Nhidd}
\path[yshift=1cm]
node[hidden neuron 2] (H2-\name) at (2\separ,-\y cm) {$\hid_{2\y}$};
\node at (2\separ,-\Nhidd+0.25) {$\vdots$};
\node[hidden neuron 2] (H2-\Nhidd+1) at (2\separ,-\Nhidd-0.75) {$\hid_{2\nhidn_2}$};



\foreach \name / \y in {1,...,\Nout}
\path[yshift=0cm]
node[output neuron] (O-\name) at (3\separ,-\y cm) {$\hat{I}_{h+\y}$};
\node at (3\separ,-\Nout-1) {$\vdots$};
\node[output neuron] (O-\Nout+1) at (3\separ,-\Nout-2) {$\hat{I}_{{h+6}}$};



\foreach \source in {1,...,\Nin}
\foreach \dest in {1,...,\Nhid}
\path (I-\source) edge (H-\dest);

\foreach \dest in {1,...,\Nhid}
\path (I-\Nin+1) edge (H-\dest);
\foreach \source in {1,...,\Nin}
\path (I-\source) edge (H-\Nhid+1);
\path (I-\Nin+1) edge (H-\Nhid+1);
\foreach \dest in {1,...,\Nhid}
\path (I-12) edge (H-\dest);
\path (I-12) edge (H-\Nhid+1);

\foreach \dest in {1,...,\Nhid}
\path (I-13) edge (H-\dest);
\path (I-13) edge (H-\Nhid+1);

\foreach \source in {1,...,\Nhid}
\foreach \dest in {1,...,\Nhidd}
\path (H-\source) edge (H2-\dest);

\foreach \dest in {1,...,\Nhidd}
\path (H-\Nhid+1) edge (H2-\dest);
\foreach \source in {1,...,\Nhid}
\path (H-\source) edge (H2-\Nhidd+1);
\path (H-\Nhid+1) edge (H2-\Nhidd+1);

%
%
%

\foreach \source in {1,...,\Nhidd}
\foreach \dest in {1,...,\Nout}
\path (H2-\source) edge (O-\dest);

\foreach \dest in {1,...,\Nout}
\path (H2-\Nhidd+1) edge (O-\dest);
\foreach \source in {1,...,\Nhidd}
\path (H2-\source) edge (O-\Nout+1);
\path (H2-\Nhidd+1) edge (O-\Nout+1);

\node[annot,above of=H-1, node distance=0.75cm] (hl) {Hidden layer};
\node[annot,above of=H2-1, node distance=0.75cm] (hl) {Hidden layer};
\node[annot,above of=I-1, node distance=0.75cm] {Input layer};
\node[annot,above of=O-1, node distance=0.75cm] {Output layer};

\end{tikzpicture}
		\caption{DNN to forecast day-ahead prices.}
		\label{fig:hiddnet}
	\end{center}
\end{figure*}

Defining by $h$ the current hour, by $\hat{\mathcal{I}}_\mathrm{E}$ the values of the ECMWF forecast, by  ${\mathcal{I}}_\mathrm{S}$ the irradiance values obtained from the satellite image, by ${\mathcal{I}}_\mathrm{c}$ the clear-sky irradiance, and by $\hat{\mathcal{I}}$ the forecasted values of the proposed model, the forecasting model can be represented as in Figure \ref{fig:hiddnet}.
\noindent In this representation, it was assumed that the optimal depth was 2 hidden layers, and that the optimal past irradiance values are lags 0, 1, and 2 w.r.t.~the current hour $h$; i.e.~$\mathcal{I}_{\mathrm{S},h}$, $\mathcal{I}_{\mathrm{S},h-1}$, $\mathcal{I}_{\mathrm{S},h-2}$; and lag 24 w.r.t.~the 6 prediction hours $h+1,\,\ldots,\,h+6$; i.e.~$\mathcal{I}_{\mathrm{S},h-23},\,\ldots,\,\mathcal{I}_{\mathrm{S},h-18}$.


\section{Case study}
\label{sec:casestudy}
In order to evaluate the proposed model, 30 sites in the Netherlands are considered and the accuracy of the proposed model is compared with that of specific models individually trained using local data.

\subsection{Data description}
The dataset spans four years, i.e.~from 01/01/2014 until 31/12/2017, and comprises, for each of the 30 sites, the following four types of input data:
\begin{enumerate}
	\item The historical ground data measured on site.
	\item The satellite-based irradiance values.
	\item The daily ECMWF forecasts.
	\item The deterministic clear-sky irradiance. 
\end{enumerate}

{In all four cases, these data represent  hourly average values between two consecutive hours. In particular, a variable given at a time step $h$ represents the average variable between hours $h$ and $h+1$, e.g.~the irradiance $\mathcal{I}_{\mathrm{S},12}$ is the average irradiance obtained from satellite images between hours 12 and 13.}

\subsubsection{Data Sources}
For the irradiance values obtained from SEVIRI satellite images, 
 the processed irradiance values are directly obtained from the \textit{Royal Netherlands Meteorological Institute (KNMI)} via their Cloud Physical Properties model \cite{knmi}.

For the ground measurements, 30 of the meteorological stations in The Netherlands that are maintained by the KNMI \cite{knmi} and that measure irradiance values using pyranometers are considered. In particular, the following 30 stations are employed: Arcen, Berkhout, Cabauw, De Kooy, De Bilt, Deelen, Eelde, Eindhoven, Ell, Gilze-Rijen, Heino, Herwijnen, Hoek van Holland, Hoogeveen, Hoorn (Terschelling), Hupsel, Lauwersoog, Leeuwarden, Lelystad, Maastricht, Marknesse,
Nieuw Beerta, Rotterdam, Schiphol, Stavoren, Twenthe,
Vlissingen, Volkel, Westdorpe, and Wijk aan Zee. \textred{The geographical location of these 30 stations is illustrated in Figure \ref{fig:siteDistribution}.}

The ECMWF forecasts are directly obtained through the ECMWF website \cite{ecmwf}. Finally, for the clear-sky irradiance, the \texttt{python} \texttt{PVLIB} library \cite{Andrews2014} that implements the clear-sky model \cite{Ineichen2002} defined in Section \ref{sec:inputs} is used.

\subsubsection{Data division}
In order to perform the study, the data is divided into three subsets:
\begin{enumerate}
	\item Training set (01/01/2014 to 31/12/2015): these 2 years of data are used for training and estimating the various models.
	\item Validation set (01/01/2016 to 31/12/2016): a year of data is used to select the optimal hyperparameters and features,  and to perform early-stopping when training the network.
	\item Test set (01/01/2017 to 31/12/2017): a year of data that is not used at any step during the model estimation process, is employed as the out-of-sample data to compare the proposed model against local models.
\end{enumerate}

\begin{figure}[htb]
	\begin{center}
	\includegraphics[width=0.97\columnwidth]{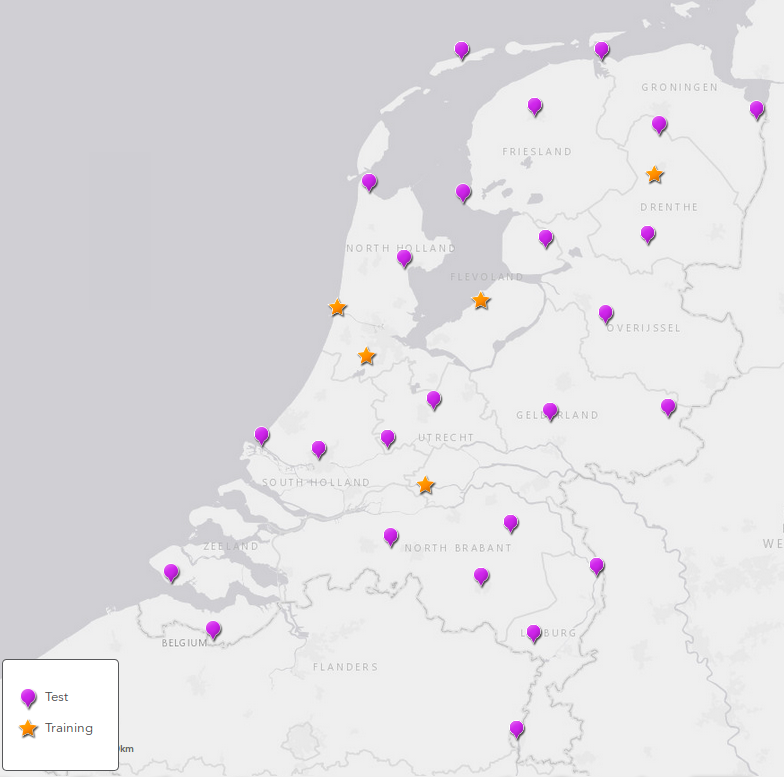}
	\end{center}
\caption{Geographical distribution of the 30 sites in the case study. The blue dots are the  5 sites used for estimating the model. The red dots represent the 25 out-of-sample sites to evaluate the model.}
\label{fig:siteDistribution}
\end{figure}

In addition to the time separation, the data is further divided according to the location:

\begin{enumerate}
	\item Of the 30 sites, 5 are used to train the proposed models. In particular, the following 5 were randomly selected: Herwijnen, Wijk aan Zee, Schiphol, Twenthe, and Lelystad.
	\item The remaining 25 act as out-of-sample data to show that the model can predict irradiance at any site without the need of local data.
\end{enumerate}
\textred{\noindent This separation is depicted in Figure \ref{fig:siteDistribution}, which represents the geographical distribution of the 30 sites distinguishing between training and test sites.}
\noindent In short, the proposed model is trained using data from 5 sites spanning three years and it is evaluated in 25 additional locations and using an additional year of data.

It is important to note that the above separation in 5+25 locations only applies for the proposed model. In particular, for the local models used as benchmark, the data division is only performed as a function of time as, by definition, each local model considers only local data.

\subsubsection{Data Preprocessing}
To evaluate the proposed models, the hours of the day for which the irradiance is very small are disregarded. In particular, those hours that correspond with solar elevation angles below $3^\circ$ are disregarded. This limitation on the solar elevation angles implies that the number of forecasts per day available to evaluate the model changes throughout the year; e.g.~while in June the model makes 11-12 forecasts per day, in January that number is reduced to 3-4. 


In addition to the above preprocessing step, the hourly time slots that have missing values are also disregarded.

\subsection{Local models}
To compare the proposed forecaster, four types of local models are considered: a persistence model \cite{Diagne2013}, an \textit{autoregressive model with exogenous inputs (ARX)} \cite{Lauret2015}, a \textit{gradient boosting tree (GBT)} algorithm \cite{Chen2016}, and a local neural network \cite{Lauret2015}. 

Moreover, in addition to the local models, it is also included in the benchmark the ECMWF forecast. By doing so, the accuracy between the time series models and the NWP forecast can be compared as a function of the prediction horizon, .

\subsubsection{Persistence model}
\label{sec:pers}
When evaluating a new model, a standard approach in the literature of irradiance forecasting  is to check whether the new model provides better predictions than a trivial model \cite{Diagne2013}. Moreover, the trivial model normally used is a persistence model, which assumes that the clear sky index $k_\mathrm{c}$ does not change from one time interval to the other \cite{Diagne2013}.

In particular, given the irradiance ${\mathcal{I}}_{h}$ at the current hour $h$, the clear sky index at $h$ is defined as the ratio of ${\mathcal{I}}_{h}$ to the clear sky irradiance ${\mathcal{I}}_{\mathrm{c},h}$, i.e.:

\begin{equation}
	k_{\mathrm{c},h} = \frac{\mathcal{I}_h}{\mathcal{I}_{\mathrm{c},h}}.
\end{equation}

\noindent Then, defining by ${\mathcal{I}}_{\mathrm{c},h+p}$ the clear sky irradiance  at the prediction time $h+p$,
%
%
the persistence model forecasts the irradiance ${\mathcal{{I}}}_{h+p}$ at the prediction time $h+p$ as follows:

\begin{equation}
{\mathcal{\hat{I}}}_{h+p} = k_{\mathrm{c},h}\,\mathcal{I}_{\mathrm{c},h+p}=\frac{\mathcal{I}_h}{\mathcal{I}_{\mathrm{c},h}}\,\mathcal{I}_{\mathrm{c},h+p}.
\end{equation}


\subsubsection{Linear model}
Another standard benchmark  choice in the literature of irradiance forecasting are autoregressive linear models \cite{Lauret2015,Diagne2013}; hence, the second model considered in the comparison is a linear autoregressive model that can optimally select its exogenous inputs. As the model is local, a different model per location, per hour of the day $h$, and for prediction time $h+p$ is considered. Therefore, as the proposed model is evaluated in 25 locations, 6 forecasts per day are made, and each forecast is made for $6$ prediction times, a total of $25\times6\times6=900$ models are estimated.

The exogenous inputs of these models are similar to the DNN, but instead of using the satellite irradiance maps $\mathcal{I}_{\mathrm{S}}$, the models consider the historical irradiance ground measurements $\mathcal{I}_{\mathrm{G}}$. In particular, the model for the prediction time $h+p$ considers the clear-sky irradiance ${\mathcal{I}}_{\mathrm{c},h+p}$ and the ECMWF forecast $\hat{\mathcal{I}}_{\mathrm{E},h+p}$ at the  prediction time. For the historical irradiance values $\mathcal{I}_{\mathrm{G}}$, as with the global model and the satellite-based irradiance $\mathcal{I}_{\mathrm{S}}$, the specific lagged values are optimally selected using the feature selection method described in Section \ref{sec:hyper}. \textred{In addition, to ensure that the differences between models are not due to differences in input data, the model is allowed to choose satellite data through the feature selection method.
}
\subsubsection{Gradient boosting tree}
As a third model, the XGBoost algorithm \cite{Chen2016} is considered, a GBT model that predicts data by combining several regression trees. In particular, the model is based on the principle of boosting \cite[Chapter 10]{Hastie2001}, i.e.~combining models with high bias and low variance in order to reduce the bias while keeping a low variance. 

{It is important to note that, while several models based on regression trees have been proposed in the literature for forecasting solar irradiance \cite{Voyant2017}, the XGBoost algorithm has, to the best of our knowledge, not yet been used. Nevertheless, including this model in the benchmark was decided for two reasons: (1) it has been shown to outperform other regression tree methods and has recently become the winner of several challenges in Kaggle, a site that hosts machine learning competitions \cite{Chen2016}; (2) it has been successfully used in other energy-based forecasting applications, e.g.~forecasting electricity prices \cite{Lago2018a}.}

As with the linear model, a different GBT per location, hour, and prediction time is estimated; i.e.~900 different models are estimated. Similarly, the model inputs are the same as the linear models, i.e.~the clear-sky irradiance ${\mathcal{I}}_{\mathrm{c},h+p}$  and the ECMWF forecast $\hat{\mathcal{I}}_{\mathrm{E},h+p}$ at the prediction time, and the historical irradiance values $\mathcal{I}_{\mathrm{G}}$ optimally selected using the feature selection method. \textred{In addition, to ensure that the differences between models are not due to differences in input data, the model is allowed to choose satellite data through the feature selection method.
}

It is important to note that, as done with the proposed DNN, all the GBT hyperparameters (see \cite{Chen2016}) are optimally selected using the hyperparameter optimization algorithm define in Section \ref{sec:hyper}.

\subsubsection{Neural network}
As a fourth model, a local DNN that considers very similar inputs, outputs, structure, and training algorithm as the proposed global DNN is considered. The main difference w.r.t.~to the proposed DNN is that it considers the local measurements of the irradiance $\mathcal{I}_{\mathrm{G}}$ in addition to the satellite irradiance maps $\mathcal{I}_{\mathrm{S}}$. However, the type and number of hyperparameters that the model optimizes are the same as for the global DNN and they are also optimized using the hyperparameter optimization algorithm defined in Section \ref{sec:hyper}.

The reason for including this model in the case study is that, similar to the linear and the persistence models, neural networks are a standard choice in the literature of solar irradiance forecasting \cite{Deneke2008,Voyant2017}.

As the proposed DNN is evaluated in 25 sites and the model is local, 25 different local DNNs are estimated. Unlike the linear and GBT models, the same DNN is used for the different hours of the day; this was done because it was empirically observed that the distinction of a different DNN per hour of the day led to worse predictive accuracy. 

\subsection{Hyperparameter Optimization and Feature Selection}
\label{sec:hyperfeat}
As defined in Section \ref{sec:modelframe}, the hyperparameters and input features of the global DNN are optimally selected according to the geographical location. {In this case study, the range of the hyperparameters considered in the optimization search and their obtained optimal values are listed in Table \ref{tab:hyper}.}

\begin{table}[h!]
\setlength{\abovecaptionskip}{-5pt}
		\caption{Optimal hyperparameters for the global DNN.}
		\label{tab:hyper}
	\small
	\renewcommand\arraystretch{1.3}
	\begin{center}
		\begin{tabular}{l c c}
			\hline
			\bfseries Hyperparameter &\bfseries  Value &\bfseries Search Range\\
			\hline			
			Number of hidden layers & 2 & \{1, 2, 3, 4\} \\
			Neurons in $1^\mr{st}$ layer & 208 & [100, 400]\\
			Neurons in $2^\mr{nd}$ layer & 63 & [50, 150]\\
			Initial Learning  Rate & $1.16\times10^{-3}$  & $[10^{-4}, 10^{-2}]$\\	
			Dropout & 0.14 & [0, 1]\\
		\end{tabular}
	\end{center}
\end{table}

\noindent In terms of the lagged satellite-based irradiance values, the optimal input features are defined by the irradiance values at lags 0, 1, 2, and 3 w.r.t.~the current hour $h$; i.e.~$\mathcal{I}_{\mathrm{S},h},\,\ldots,\,\mathcal{I}_{\mathrm{S},h-3}$; and at lag 24 w.r.t~the 6 prediction hours $h+1,\,\ldots,\,h+6$; i.e.~$\mathcal{I}_{\mathrm{S},h-23},\,\ldots,\,\mathcal{I}_{\mathrm{S},h-18}$.

For the local models, the hyperparameters and input features are also optimized. However, considering that 900 linear models, 900 GBT models, and 25 local DNNs are used, displaying all their optimal hyperparameters and input features is out of the scope of this paper. However, the main results can be summarized as follows:

\begin{enumerate}
\textred{	\item In terms of input features, all the local models (except for the persistence model) performed two types of selection:
	\begin{enumerate}
		\item Use satellite data in addition to local data.
		\item Choose the relevant historical irradiance values.
	\end{enumerate} 
	The addition of satellite data did not improve the performance w.r.t.~using ground data only; therefore, none of the local models considered this information. In addition, in terms of ground irradiance values $\mathcal{I}_{\mathrm{G}}$, all the local models consider the irradiance values at lags 0 and 1 w.r.t.~the current hour $h$ and at lag 24 w.r.t.~the prediction hour $h+p$. In addition, most of them also consider the irradiance values at lags 2 and 3 w.r.t.~the current hour $h$; the exception are models that predict the solar irradiance at early hours of the day when lags of 2-3 hours represent irradiance values of 0.}
	\item In the case of the local DNNs, the number of hidden layers is 2 for all 25 sites. Moreover, the number of neurons in the first (second) hidden layer varies from 95 to 242 (51 to 199) neurons depending on the site. Similarly, the dropout and the learning rate respectively oscillate between 0 and 0.45, and between $5.825\times10^{-4}$ and $5.800\times10^{-2}$.
	\item In the case of the GBT models, the range of the hyperparameters values varies in a larger range, e.g.~the number of trees per model fluctuates between 10 and 1000 and the depth of each tree varies between 1 and 20.
\end{enumerate}

\textred{\subsection{Overall results}
After defining the setup of the case study and describing the selection of hyperparameters and features, in this section the average performance of the global DNN is compared against that of the local models. Particularly, the first metrics to take into account to compare the models are the average metrics; i.e.~rRMSE, forecasting skill $s$, and MBE; across the 25 sites and the 6 prediction times. These average metrics are listed in Table \ref{tab:gencom}, where the forecasting skill was computed using the same window length employed in \cite{Marquez2012}, i.e.~200 samples\footnote{As in \cite{Marquez2012}, the window length for which $s$ was stable was analyzed. Similar to \cite{Marquez2012}, 200 samples were found to be a reasonable value.}. 

\begin{table}[h!]
	\setlength{\abovecaptionskip}{-5pt}
	\caption{Comparison of the average predictive accuracy across sites and prediction times by means of rRMSE, forecasting skill  $s$, and MBE. }
	\label{tab:gencom}
	\small
	\renewcommand\arraystretch{1.3}
	\begin{center}
		\begin{tabular}{l c c c}
			\hline
			\bfseries Model &\bfseries  rRMSE [\%] & \bfseries s [\%] & \bfseries MBE [W/m$^2$] \\
			\hline			
			Global DNN & 31.31 &  22.42 & -1.04\\
			Linear & 32.01 & 21.22 & -1.07\\
			Local DNN &  32.10 & 19.29 & -1.43\\
			ECMWF&	34.94 & 9.75 & -2.52\\			
			GBT & 35.85  & 9.92 & 1.50\\	
			Persistence & 41.98 & 0 & 11.60\\
		\end{tabular}
	\end{center}
\end{table}

From Table \ref{tab:gencom}, several observations can be drawn:

\begin{enumerate}
	\item In terms of square errors, i.e.~rRMSE, the predictive accuracy of the proposed global model is  slightly better than all the local models and significantly better than some of them,  in particular the GBT model or the persistence model. Among the local models, both the linear and local DNN perform the best and the persistence model the worst.
		\item This same observation can be inferred from looking at the forecasting skill: the proposed global model performs similar to the linear model, slightly better than the local DNN, and much better than the other models. In addition, when compared across all sites and predictive horizons, all models perform better than the persistence model.
\item In terms of model bias, i.e.~MBE, all models show a very small bias that indicates that the models are not biased. Particularly, considering that the average irradiance of the dataset is approximate 350\,W/m$^2$, the bias of all the models is around 0.3-0.8\% of the average irradiance, which represents a negligible bias. The exception to this is the persistence model, whose bias of  3\% of the average irradiance is a bit larger, but still quite small.
\end{enumerate}
}
\textred{\subsection{Comparison with previously validated forecast models}
While the proposed global model seems to be a good replacement of the local models considered in this paper,  it is also very important to establish its quality w.r.t.~previously validated forecast models from the literature. As explained in Section \ref{sec:metrics}, while this comparison cannot fairly be done using a metric like rRMSE, it can be roughly assessed using the forecasting skill $s$. In particular, using the results of \cite{Marquez2012}, we can establish a comparison between the proposed global model, the local NARX model proposed in \cite{Marquez2012}, and the cloud motion forecast of \cite{Perez2010}. As both models from the literature were originally only evaluated for 1-hour step ahead forecasts, we also limit the comparison of the global model to that interval. The comparison is listed in Table \ref{tab:complit}.

\begin{table}[h!]
	\setlength{\abovecaptionskip}{-5pt}
	\caption{Comparison of the average predictive accuracy between the global model, a NARX model from the literature, and a cloud moving forecast from the literature. The comparison is done for 1-hour ahead forecasts and by means of forecasting skill .}
	\label{tab:complit}
	\small
	\renewcommand\arraystretch{1.3}
	\begin{center}
		\begin{tabular}{l c}
			\hline
			\bfseries Model & \bfseries s [\%] \\
			\hline			
			Global DNN & 10 \\
			NARX \cite{Marquez2012} & 12 \\
			Cloud moving \cite{Perez2010} &  8 \\
		\end{tabular}
	\end{center}
\end{table}

What can be observed from these results is that the overall quality of the proposed global model for 1-hour ahead forecasts is very similar to those from the literature. Therefore, as initially observed when comparing the average performance of the global model w.r.t.~to the local model considered in this paper, the proposed global model seems to be an excellent candidate to save the operational costs of installing local sensors and collecting ground measurements.
}
\textred{
\subsection{Comparison across prediction horizons}
A third step required to analyze the performance of the proposed global model is to verify that its average performance is satisfied across all prediction times. In particular, it is important to check whether the global models can build accurate predictions at all short-term horizons. To perform this comparison, the two metrics used for comparing predictive accuracy, i.e.~rRMSE and the forecasting skill $s$, are evaluated for each benchmark model and predictive horizon. This comparison is listed in Table \ref{tab:combySA} and illustrated in Figure \ref{fig:BySA}.

\begin{table}[h!]
		\setlength{\abovecaptionskip}{-5pt}
	\caption{Comparison of the predictive accuracy of the various forecasters across the 6 prediction times by means of rRMSE and forecasting skill $s$. The best model is marked with bold font.}
\label{tab:combySA}		
	\small
		\setlength{\tabcolsep}{4pt}
	\renewcommand\arraystretch{1.3}
	\begin{center}
		\begin{tabular}{l| c c c c c c}
			\bfseries  Horizon [h]& 1 & 2  & 3 &4 &5 &6 \\
			\hline
			\bfseries Model &\multicolumn{6}{c}{  rRMSE [\%]}\\
			\hline			
			Global DNN &   \bfseries  25.07& \bfseries30.18& \bfseries32.36& \bfseries34.19& \bfseries36.10& 38.71\\
			Linear &    26.67&  31.36&  33.11&  34.63&  36.44&  \bfseries 38.35\\
			Local DNN &   26.82 & 30.90 & 32.91 & 34.67 & 36.68 & 39.88 \\	
			 GBT   &   30.05&  34.78&  36.95&  39.04&  40.67&  43.59 \\
			Persistence  &  28.74&  36.89&  42.29&  47.28&  52.05&  56.69\\
			ECMWF &   35.91&  35.01&  35.12&  35.91&  37.45&  39.28 \\
			\hline
			\bfseries  &\multicolumn{6}{c}{  $s$ [\%]}\\
\hline			
Global DNN &  \bfseries 9.98 & \bfseries18.38 & \bfseries23.40 & \bfseries27.04 & \bfseries 28.30 & 27.38 \\
Linear &   7.67 &   15.71 &   21.73 &   26.03 &   27.76 &   \bfseries28.42 \\
Local DNN &     6.34 &  16.98 &  22.13 &  22.64 &  25.13 &  22.51 \\	
GBT   &  -5.18 &   6.06 &   12.23 &   15.29 &   16.00 &   15.11 \\
Persistence  &  0&  0&  0&  0&  0&  0\\
ECMWF &     -29.07 &   4.68 &   16.77 &   22.74 &   23.23 &   20.19  \\			
		\end{tabular}
	\end{center}
\end{table}

\setlength{\figW}{0.49\textwidth}
\setlength{\figH}{0.66\figW}
\begin{figure}[htb]
	\begin{center}
	\begin{subfigure}{\columnwidth}
\begin{tikzpicture}

\definecolor{color1}{rgb}{1,0.498039215686275,0.0549019607843137}
\definecolor{color0}{rgb}{0.12156862745098,0.466666666666667,0.705882352941177}
\definecolor{color3}{rgb}{0.83921568627451,0.152941176470588,0.156862745098039}
\definecolor{color2}{rgb}{0.172549019607843,0.627450980392157,0.172549019607843}
\definecolor{color4}{rgb}{0.580392156862745,0.403921568627451,0.741176470588235}
\definecolor{color5}{rgb}{0.549019607843137,0.337254901960784,0.294117647058824}

\begin{axis}[
xlabel={Prediction Horizon [h]},
ylabel={rRMSE [\%]},
xmin=0.75, xmax=6.25,
ymin=23.487624201196, ymax=59.3034534725948,
width=\figW,
height=\figH,
tick align=outside,
tick pos=left,
x grid style={lightgray!92.02614379084967!black},
y grid style={lightgray!92.02614379084967!black},
legend style={at={(0.03,0.97)}, anchor=north west, draw=white!80.0!black,font=\small},
legend columns=3, 
legend cell align={left},
legend entries={{Global},{Linear},{Local DNN},{GBT},{Persistence},{ECMWF}}
]
\addplot [ultra thick,black, mark=*, mark size=2, mark options={solid}]
table {%
1 25.0787711880354
2 30.1827693397984
3 32.360913575472
4 34.1970869559918
5 36.1091968568801
6 38.7153129188762
};
\addplot [thick,dash pattern=on 2pt off 5pt on 5pt off 5pt,color3, mark=square*, mark size=2, mark options={solid}]
table {%
1 26.6740187839708
2 31.3650759274759
3 33.1256184236544
4 34.6394829571378
5 36.4456128470152
6 38.3564969605087
};
\addplot [very thick,densely dotted,color5, mark=triangle*, mark size=3, mark options={solid}]
table {%
1 26.8202224822728
2 30.903664778619
3 32.9196603343957
4 34.6711371251352
5 36.68961448692
6 38.3896794802773
};
\addplot [very thick,loosely dotted, color0, mark=diamond*, mark size=3, mark options={solid}]
table {%
1		30.0541
2		34.7857
3		36.9542
4		39.0421
5		40.6702
6		43.5963
};
\addplot [thick, dash pattern=on 1pt off 3pt on 3pt off 3pt,color2, mark=pentagon*, mark size=2, mark options={solid}]
table {%
1		28.7435
2		36.8947
3		42.2958
4		47.2856
5		52.0555
6		56.6947
};
\addplot [very thick,dashed, color4, mark=star, mark size=4, mark options={solid}]
table {%
1		35.9103
2		35.0197
3		35.1246
4		35.9106
5		37.4541
6		39.2885

};
\end{axis}

\end{tikzpicture}
	\caption{Comparison by means of rRMSE.}
\end{subfigure} 	
\begin{subfigure}{\columnwidth}
\begin{tikzpicture}
\definecolor{color1}{rgb}{1,0.498039215686275,0.0549019607843137}
\definecolor{color0}{rgb}{0.12156862745098,0.466666666666667,0.705882352941177}
\definecolor{color3}{rgb}{0.83921568627451,0.152941176470588,0.156862745098039}
\definecolor{color2}{rgb}{0.172549019607843,0.627450980392157,0.172549019607843}
\definecolor{color4}{rgb}{0.580392156862745,0.403921568627451,0.741176470588235}
\definecolor{color5}{rgb}{0.549019607843137,0.337254901960784,0.294117647058824}

\begin{axis}[
xlabel={Prediction Horizon [h]},
ylabel={FSkill [\%]},
xmin=0.75, xmax=6.25,
ymin=-31.9485058482401, ymax=31.2968347221308,
width=\figW,
height=\figH,
tick align=outside,
tick pos=left,
 y label style={at={(-0.097,0.5)}},
x grid style={lightgray!92.02614379084967!black},
y grid style={lightgray!92.02614379084967!black},
legend style={at={(0.99,0.03)}, anchor=south east, draw=white!80.0!black,font=\small},
legend columns=3, 
legend cell align={left},
legend entries={{Global},{Linear},{Local DNN},{GBT},{Persistent},{ECMWF}},
]
\addplot [ultra thick,black, mark=*, mark size=2, mark options={solid}]
table {%
1 9.98962764924296
2 18.3843160113606
3 23.4016709705138
4 27.0445946047915
5 28.3004543441869
6 27.3825224952807
};
\addplot [thick,dash pattern=on 2pt off 5pt on 5pt off 5pt,color3, mark=square*, mark size=2, mark options={solid}]
table {%
	1 7.6787353111269
	2 15.7140896769726
	3 21.7399508012991
	4 26.0362413565727
	5 27.7683679211933
	6 28.4220465143866
};
\addplot [very thick,densely dotted,color5, mark=triangle*, mark size=3, mark options={solid}]
table {%
	1 6.34083490686788
	2 16.9872635342824
	3 22.1320199114686
	4 22.6479974148521
	5 25.1380728746863
	6 22.5172835980867
};
\addplot [very thick,loosely dotted, color0, mark=diamond*, mark size=3, mark options={solid}]
table {%
	1 -5.18217733042643
	2 6.06145920296365
	3 12.2387473659874
	4 15.291539616356
	5 16.0000487980459
	6 15.1143274211404
};

\addplot [thick, dash pattern=on 1pt off 3pt on 3pt off 3pt,color2, mark=pentagon*, mark size=2, mark options={solid}]
table {%
1 6.69011792631125e-08
2 2.27978521616379e-08
3 3.81583479258651e-08
4 1.41682844345681e-08
5 2.9777648791196e-08
6 3.05918798864419e-08
};

\addplot [very thick,dashed, color4, mark=star, mark size=4, mark options={solid}]
table {%
1 -29.073717640496
2 4.68701912118344
3 16.7755115931973
4 22.7438117299841
5 23.2308000036987
6 20.1911983366746
};
\end{axis}

\end{tikzpicture}
	\caption{Comparison by means of the forecasting skill $s$}	
\end{subfigure} 		
		\caption{Comparison of the predictive accuracy of the various forecasters across the 6 prediction times.}
		\label{fig:BySA}
	\end{center}
\end{figure}
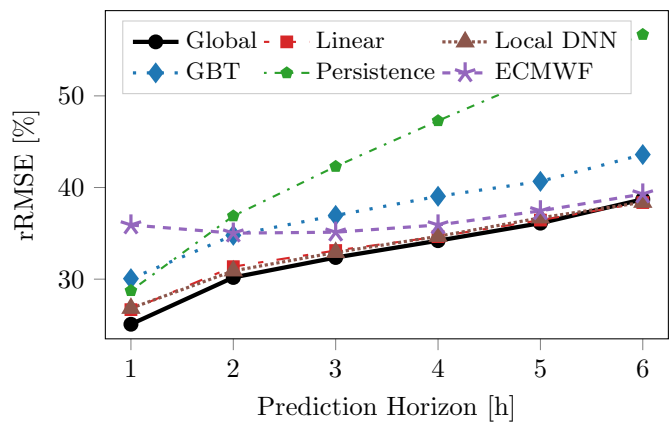
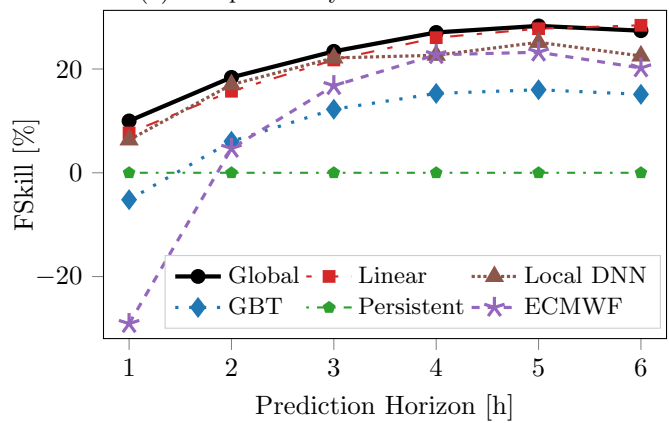

As can be seen from Table \ref{tab:combySA} and Figure \ref{fig:BySA}, the global model seems to be the best model for the first 5 prediction horizons (both in terms of rRMSE and forecasting skill $s$), and the second best (very close to the best one) for the last prediction horizon. Based on these results it can be observed that not only the global model is overall equal or better than the local models, but it also performs equally well or better than them across all prediction horizons. As a result, the proposed model is a very promising candidate to replace the local models and to save operational costs without compromising the forecasting quality.

In addition to this analysis of the global model performance, three additionally interesting observations can be made:

\begin{enumerate}
	\item The persistence model is the worst across all prediction horizons except the first one. This result agrees with previous results from the literature \cite{Diagne2013} that stated that the persistence model only provides reasonable results for prediction horizons shorter than 1 hour.
	\item Among the local models, the linear and DNN models show the best performance across all 6 prediction horizons.
	\item The ECMWF forecast improves its accuracy relatively to the other models as the prediction time increases. In particular, in the case of the last prediction time, the ECMWF forecast has almost the same performance as the global DNN and the linear models. Considering previous results from the literature \cite{Diagne2013}, this is highly expected as NWP models start to perform better than time series models for prediction horizons larger than 4-6 hours.
	\item For 1 hour ahead predictions, the ECMWF model is the worst; specially, considering its $s$ value for the first prediction horizon, the weather-based model is much worse than a simple persistence model.
\end{enumerate}
}
\subsection{Comparison across geographical site}
The final step to analyze the better or equal performance of the global model is to validate whether the quality of the performance is kept across the 25 different sites. In particular, it is important to check whether the global model can generalize and build accurate predictions across all geographical locations. For the same of simplicity, this comparison is only done in terms of the rRMSE metric; in particular, as it was the case with all previous results, the values of the forecasting skill $s$ fully agree with the rRMSE across all locations, and they are a bit redundant. The comparison across the geographical locations is listed in Table \ref{tab:combySite} and illustrated in Figure \ref{fig:BySite}.

\newcommand{\cgreen}{\cellcolor{greenPlots!60!white}}
\newcommand{\cred}{\cellcolor{redPlots!60!white}}
\begin{table*}[h]
			\setlength{\abovecaptionskip}{-5pt}
	\caption{Comparison of the predictive accuracy of the various forecasters by means of rRMSE. The best model is marked with bold font.}
\label{tab:combySite}			
	\footnotesize
	\setlength{\tabcolsep}{3pt}
	\renewcommand\arraystretch{1.3}
	\begin{center}
		\begin{tabular}{l| c c c c c c c c c c c c c}
			&\multicolumn{13}{c}{\bfseries Site}\\
			\hline
			\bfseries Model & 
  Arcen &  Berkhout &  Cabauw &  De Kooy & Lauwers.  &  Deelen & Maastric.  &  Eindhov. &  Westdorpe &  Gilze-R. &  Heino &  Hoek v.~H.&Ell \\
			\hline			
			Global &\bfseries  32.39& \bfseries30.24& \bfseries30.75& \bfseries29.49&\bfseries30.32&\bfseries34.55&\bfseries30.82&32.11&32.07&\bfseries 32.37&32.80&\bfseries29.24&\bfseries32.42 \\
							
			Linear &  33.03&31.05&31.01& 29.87&31.16&35.47&31.73&32.28&33.11& 32.89&\bfseries32.75&30.53&32.50 \\

			 DNN &33.43&32.77&31.27& 31.14&30.95&35.75&31.48&\bfseries32.03&\bfseries31.93& 33.04&32.89&29.66&32.77 \\
			 			
 GBT   & 35.80&35.41&35.20& 33.45&35.79&39.62&35.68&37.22&36.33& 36.30&36.88&33.61&37.35 \\
 
Persistence  &43.63 &41.04 &41.51 & 41.18 &41.14 &45.47 &41.20 &43.20 &40.28 & 42.86 &43.80 &40.59 &42.65
\\
ECMWF & 35.21&34.09&33.95& 32.94&33.83&38.61&34.93&34.95&36.63& 35.73&35.32&33.39&36.12
 \\	
	\hline
	\hline
\bfseries Model &Hoorn  &  Hoogev.  &  Hupsel & De Bilt  &  Leeuward. &  Eelde  &  Marknes.  &  Rotterd. &  Stavoren &  Vlissing. &  Volkel &    Nieuw B.\\
 			\hline			
 Global & \bfseries
  29.63&\bfseries31.44&32.88&\bfseries31.68&\bfseries30.16&\bfseries31.58&\bfseries31.19&\bfseries30.21&\bfseries29.38&\bfseries30.81&\bfseries32.46&32.37 \\
  
 Linear & 30.63&32.05&\bfseries32.82&32.11&30.51&32.20&31.30&31.54&30.51&32.23&33.04&33.52 \\
 
   DNN & 30.24&33.05&32.83&32.02&31.97&31.62&31.72&31.25&29.85&31.92&34.68&\bfseries32.34 \\	
   
 GBT   & 34.46&36.44&36.99&35.94&35.20&36.19&35.30&34.53&34.14&35.50&36.50&36.98\\
 
 Persistence  & 40.35  &42.51  &42.42  &43.61  &40.80  &42.04  &41.24  &40.92  &40.01  &41.11  &42.71  &43.58 \\		

ECMWF & 33.62&35.19&35.11&34.69&34.17&35.34&34.85&34.92&33.35&35.05&35.33& 36.27 \\
\end{tabular}
	\end{center}
\end{table*}

\setlength{\figW}{\textwidth}
\setlength{\figH}{0.4\figW}
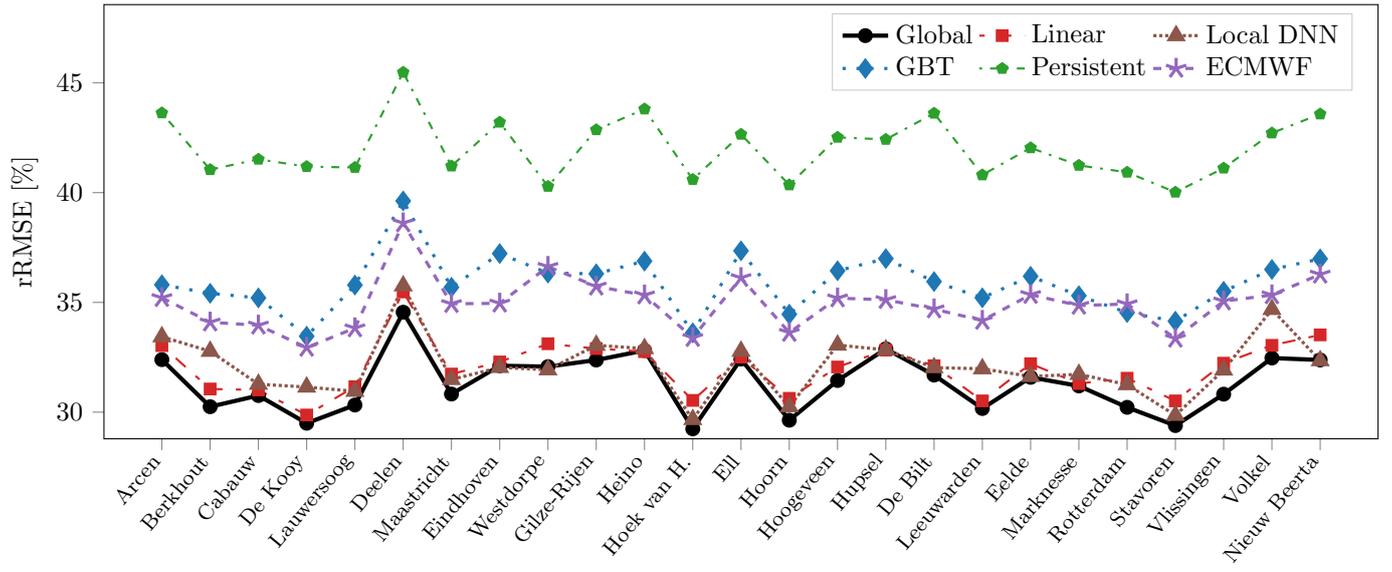
\begin{figure*}[!htb]
	\begin{center}
\begin{tikzpicture}

\definecolor{color1}{rgb}{1,0.498039215686275,0.0549019607843137}
\definecolor{color0}{rgb}{0.12156862745098,0.466666666666667,0.705882352941177}
\definecolor{color3}{rgb}{0.83921568627451,0.152941176470588,0.156862745098039}
\definecolor{color2}{rgb}{0.172549019607843,0.627450980392157,0.172549019607843}
\definecolor{color4}{rgb}{0.580392156862745,0.403921568627451,0.741176470588235}
\definecolor{color5}{rgb}{0.549019607843137,0.337254901960784,0.294117647058824}

\begin{axis}[
ylabel={rRMSE [\%]},
xmin=-0.2, xmax=26.2,
ymin=28.7865164631992, ymax=48.5655355762798,
width=\figW,
height=\figH,
xtick={1,2,3,4,5,6,7,8,9,10,11,12,13,14,15,16,17,18,19,20,21,22,23,24,25},
xticklabels={Arcen,Berkhout,Cabauw,De Kooy,Lauwersoog,Deelen,Maastricht,Eindhoven,Westdorpe,Gilze-Rijen,Heino,Hoek van H., Ell,Hoorn ,Hoogeveen,Hupsel,De Bilt,Leeuwarden,Eelde,Marknesse,Rotterdam,Stavoren,Vlissingen,Volkel,Nieuw Beerta},
tick align=outside,
tick pos=left,
x grid style={lightgray!92.02614379084967!black},
y grid style={lightgray!92.02614379084967!black},
x tick label style={rotate=50,anchor=east,font=\footnotesize},
legend columns=3, 
legend style={draw=white!80.0!black},
legend cell align={left},
legend entries={{Global},{Linear},{Local DNN},{GBT},{Persistent},{ECMWF}}
]
\addplot [ultra thick, black, mark=*, mark size=2, mark options={solid}]
table {%
1		 32.39350813800939
2		 30.24852804915662
3		 30.75664686471173
4		 29.495523922809413
5		 30.32486238817145
6		 34.55307091635774
7		 30.82566483021934
8		 32.114177071721306
9		 32.07481076614112
10		 32.374442631760514
11		 32.809605024742167
12		 29.239973523322693
13		 32.420467954377535
14		 29.63349694906824
15		 31.442253798832964
16		 32.88556442375681
17		 31.68844889483501
18		 30.16601099585474
19		 31.587713496134767
20		 31.19461799817377
21		 30.219756311185675
22		 29.388918548036125
23		 30.81968000149397
24		 32.46618221199626
25		 32.37733727728638
};
\addplot [thick,dash pattern=on 2pt off 5pt on 5pt off 5pt,color3, mark=square*, mark size=2, mark options={solid}]
table {%
1 		33.030340785157636
2 		31.0555332003417
3 		31.01191980793483
4 		29.870395532785005
5 		31.16252562382317
6 		35.47609199758493
7 		31.73618302323825
8 		32.287606798801094
9 		33.11682080735022
10 		32.89624769183935
11 		32.75711286781955
12 		30.535685386601136
13 		32.50825769681696
14 		30.63033676587317
15 		32.05747055696891
16 		32.82898866645629
17 		32.11509976569661
18 		30.513839065824516
19 		32.206981088107656
20 		31.303139299747645
21 		31.54621008117452
22 		30.512782447618614
23 		32.23170264748242
24 		33.04453202978526
25 		33.5204699465716
};
\addplot [very thick,densely dotted,color5, mark=triangle*, mark size=3, mark options={solid}]
table {%
1		 33.43614321981537
2		 32.7734593768064
3		 31.274643109413536
4		 31.14471014801313
5		 30.95095962965874
6		 35.753406892140693
7		 31.4867662497329
8		 32.037686107286295
9		 31.93079261248584
10		 33.046170033633804
11		 32.896898237838457
12		 29.6653332749215
13		 32.77596292993063
14		 30.249887532621944
15		 33.056957004626286
16		 32.838486635777386
17		 32.02819811308911
18		 31.977449942800945
19		 31.62612527861246
20		 31.726244151870353
21		 31.253320596881956
22		 29.855480475874385
23		 31.920684775718894
24		 34.68070238110743
25		 32.345541383487736
};

\addplot [very thick,loosely dotted, color0, mark=diamond*, mark size=3, mark options={solid}]
table {%
1		 35.803746535509107
2		 35.413849062758923
3		 35.20556421768935
4		 33.45986273470281
5		 35.791056103706875
6		 39.62209736213841
7		 35.689009011893025
8		 37.22494233957855
9		 36.335829344873083
10		 36.30320305409685
11		 36.881297581554673
12		 33.61562598069674
13		 37.35027378968581
14		 34.462550667719255
15		 36.442413775066923
16		 36.995591337198824
17		 35.949007539028405
18		 35.209289809982286
19		 36.190199924673716
20		 35.305490328783135
21		 34.53616880081832
22		 34.14550945234464
23		 35.50567186747768
24		 36.501261175459543
25		 36.983856336269166
};
\addplot [thick, dash pattern=on 1pt off 3pt on 3pt off 3pt,color2, mark=pentagon*, mark size=2, mark options={solid}]
table {%
1		 43.63101226876429
2		 41.04751876177622
3		 41.519038120326657
4		 41.18578771882768
5		 41.14521290379025
6		 45.479433339746633
7		 41.2090162805999
8		 43.20712631791617
9		 40.28321708818565
10		 42.86231714632314
11		 43.808624183674083
12		 40.597923483544796
13		 42.65545646039609
14		 40.35121312730309
15		 42.51846679392763
16		 42.42389825853899
17		 43.614775379317
18		 40.80536247996782
19		 42.040572258977715
20		 41.240856825269595
21		 40.927708367883425
22		 40.019194767296506
23		 41.11650071856265
24		 42.71531691879944
25		 43.580864327672325
};

\addplot [very thick,dashed, color4, mark=star, mark size=4, mark options={solid}]
table {%
1 		35.210368674373715
2 		34.095899780517824
3 		33.956739300673566
4 		32.943907282651513
5 		33.83130416537636
6 		38.61321667850174
7 		34.93730404315964
8 		34.95839437769224
9 		36.630480396174786
10 		35.731463799469976
11 		35.329205391509694
12 		33.39036567850497
13 		36.121448949229007
14 		33.62400471271314
15 		35.19430429547684
16 		35.11952499915982
17 		34.695154099248343
18 		34.17303955492765
19 		35.349917856839597
20 		34.85697861920689
21 		34.927101303595054
22 		33.35779630246242
23 		35.05509805059874
24 		35.335942804811754
25 		36.276915819716465
};
\end{axis}
\end{tikzpicture}
		\caption{Comparison of the predictive accuracy of the various forecasters across the 25 locations.}
		\label{fig:BySite}
	\end{center}
\end{figure*}

As it can be seen from Table \ref{tab:combySite} and Figure \ref{fig:BySite}, the global model seems to validate and maintain its performance across all geographical locations. In particular, analyzing this results, it is clear that the global model performs equal or better than the local models across all 25 sites. In particular, as listed in Table \ref{tab:combySite}, the global DNN is the best model for 20 of the 25 locations, and shows an rRMSE performance that is very similar to the best model in the remaining 5 locations. \textred{Therefore, it can again be conclude that the global model is a good replacement for the local models as the performance of the former is, at least, equal to the performance of the latter.}

\subsubsection{Geographical dependences}

\begin{figure*}[h!]
	\begin{subfigure}{0.33\textwidth}
		\includegraphics[width=\columnwidth]{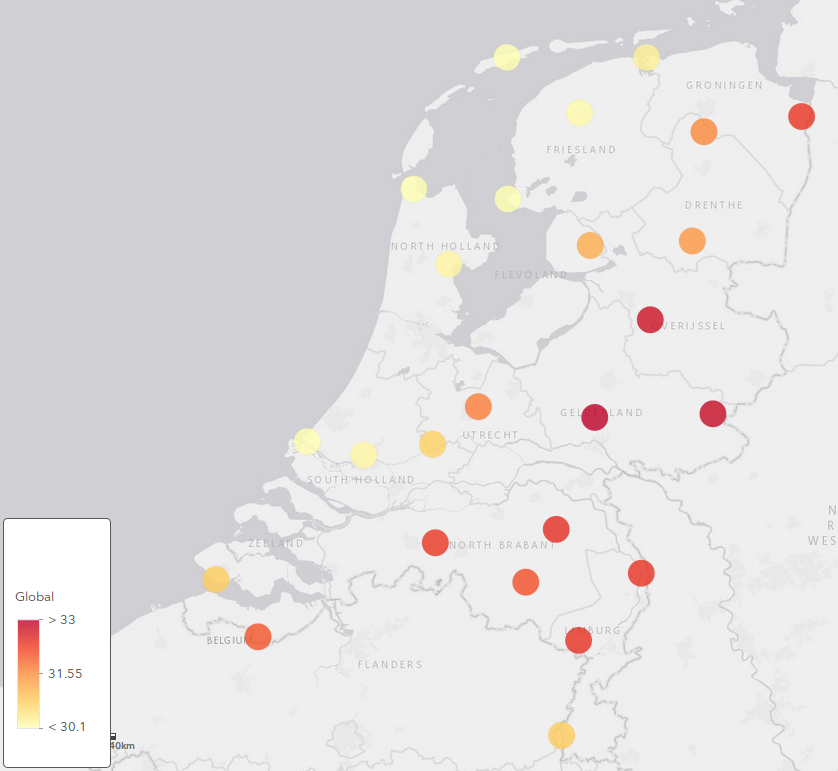}
		\caption{Global}
	\end{subfigure} 	
	\begin{subfigure}{0.33\textwidth}
	\includegraphics[width=\columnwidth]{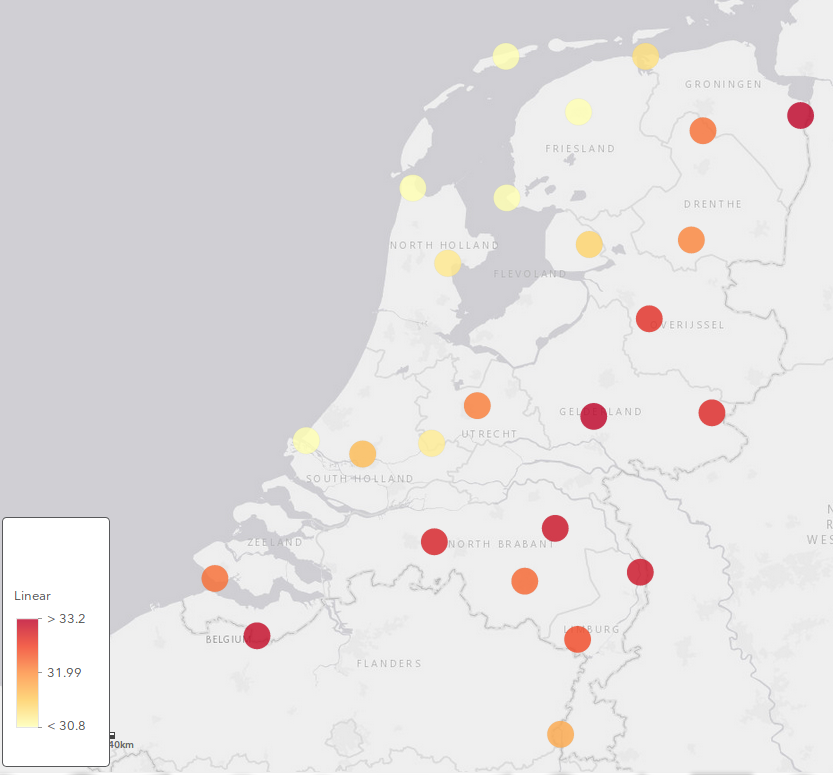}
	\caption{Linear}
\end{subfigure} 	
	\begin{subfigure}{0.33\textwidth}
	\includegraphics[width=\columnwidth]{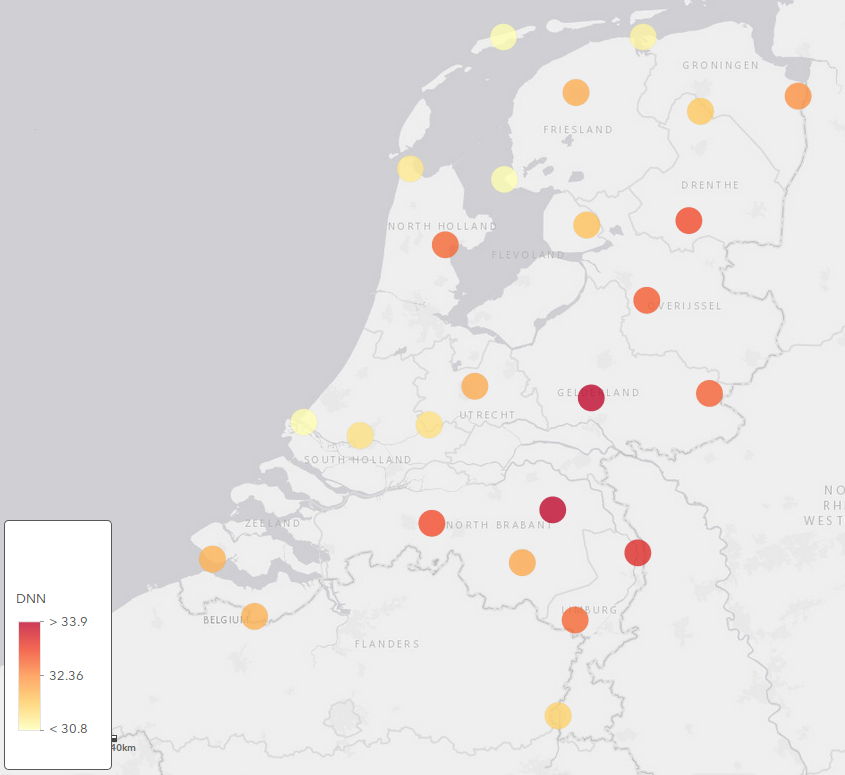}
	\caption{Local DNN}
\end{subfigure} 	
	\begin{subfigure}{0.33\textwidth}
	\includegraphics[width=\columnwidth]{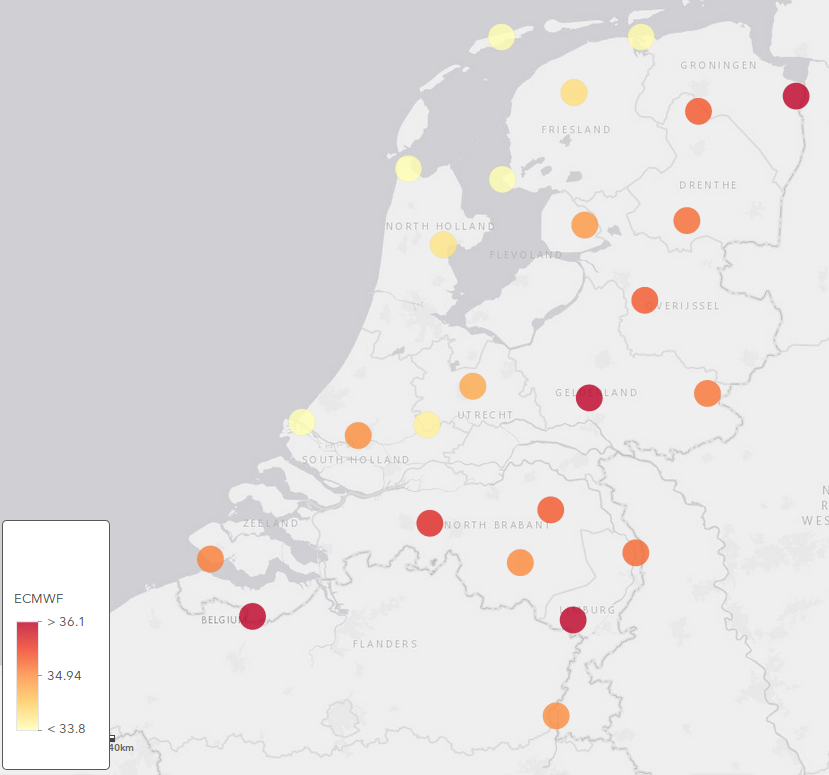}
	\caption{ECMWF}
\end{subfigure} 	
	\begin{subfigure}{0.33\textwidth}
	\includegraphics[width=\columnwidth]{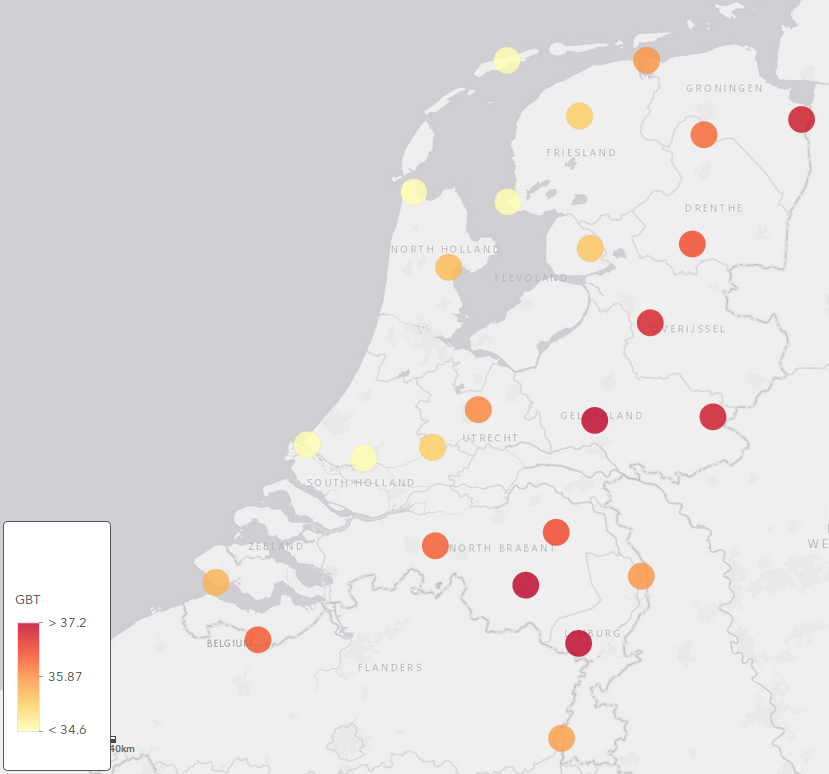}
	\caption{GBT}
\end{subfigure} 	
	\begin{subfigure}{0.33\textwidth}
	\includegraphics[width=\columnwidth]{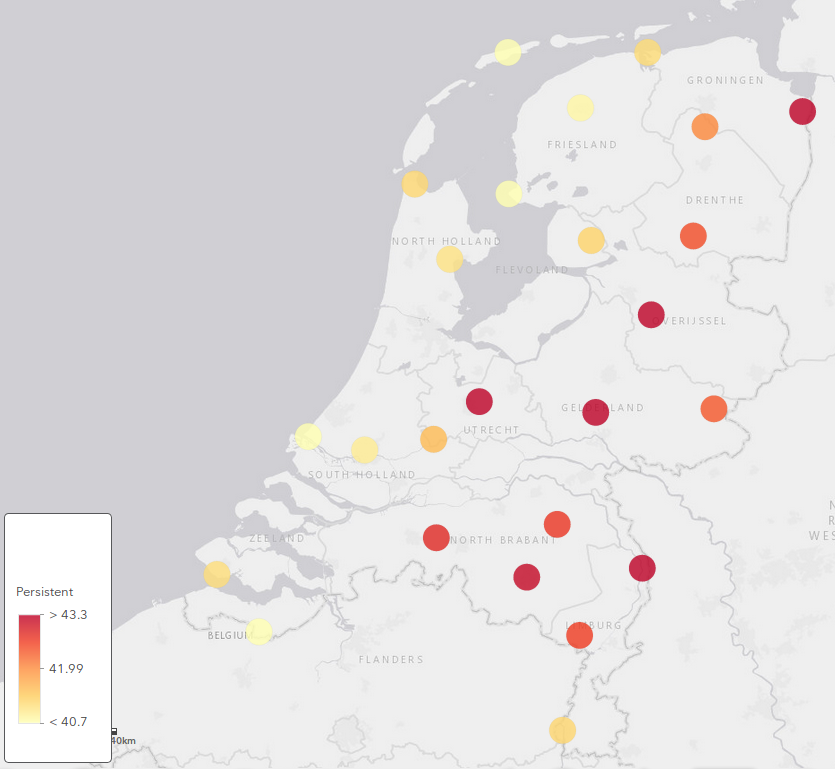}
	\caption{Persistent}
\end{subfigure} 	
	\caption{Geographical distribution of the rRMSE based on the 25 out-of-sample sites. Across the 6 models, it can be observed a clear difference between inland locations and coastal locations, with the latter having lower rRMSEs.}
\label{fig:siteDistribution2}
\end{figure*}

\textred{An interesting study to analyze is whether the rRMSE has any geographical dependence, i.e.~it might be possible that geography or climate might have an effect on the rRMSE. To study this effect, a color map with the geographical distribution of the rRMSE can be used. Such a plot is represented in Figure \ref{fig:siteDistribution2}, which depicts the geographical distribution of the rRMSE for the 6 different models. As can be observed, there is a clear difference between coastal and island sites with the latter displaying rRMSEs that are consistently higher. While this difference is not notorious, it does seem to indicate that forecasting solar irradiance at inland locations is slightly harder than at coastal sites. While analyzing the causality behind this difference is out of the scope of this paper, it is worth noting possible reasons that might cause it; particularly, differences in climate, altitude, or simple differences in irradiance ranges might explain this effect.}

\subsubsection{rRMSE distribution}
\textred{A second interesting study is to analyze the rRMSE distribution across sites. In particular, while the variability of the rRMSE can be visually observed in Figure \ref{fig:BySite}, it is interesting to analyze its empirical distribution. To  perform this analysis, the histogram of the rRMSE across the 25 sites is built for each of the 6 models. This is depicted in Figure \ref{fig:BySiteDist}, where each histogram bin represents a width of $0.5\%$ rRMSE. As it can be observed, the rRMSE distribution across the 6 locations is very similar with an interval spanning a width of 3\%-4\% rRMSE where the distribution is quite homogeneous and uniform, and an outlier on the right side representing a location with a much worse rRMSE. As can be seen from Figures \ref{fig:BySite} and \ref{fig:siteDistribution2}, this site representing the worst case-scenario is the same for all models: Deelen. Based on this result it can be concluded that, while the rRMSE is site-dependent, the range of variability of the rRMSE is small.
}

\setlength{\figW}{0.33\textwidth}
\setlength{\figH}{\figW}
\begin{figure*}[h!]
	\begin{subfigure}{0.32\textwidth}
		\begin{tikzpicture}
\begin{axis}[
    ybar,
ymin=0,
ymax=8,
xmin=29,
xmax=35,
    xtick={29,30,31,32,33,34,35},
    tick align=outside,
    tick pos=left,
    x grid style={lightgray!92.02614379084967!black},
    y grid style={lightgray!92.02614379084967!black},
    xlabel={rRMSE [\%]},
    ylabel={Number of occurrences},
    width=\figW,
    height=\figH,
]
\addplot +[
    hist={
        bins=12,
        data min=29,
        data max=35
    }
] table [y index=0] {data.csv};
\end{axis}
\end{tikzpicture}
		\caption{Global}
	\end{subfigure} \hfill
	\begin{subfigure}{0.32\textwidth}
		\begin{tikzpicture}
\begin{axis}[
    ybar,
ymin=0,
ymax=8,
xmin=29.5,
xmax=35.5,
    xtick={29,30,31,32,33,34,35,36},
    tick align=outside,
    tick pos=left,
    x grid style={lightgray!92.02614379084967!black},
    y grid style={lightgray!92.02614379084967!black},
    xlabel={rRMSE [\%]},
        width=\figW,
    height=\figH,
]
\addplot +[
    hist={
        bins=12,
        data min=29.5,
        data max=35.5
    }
] table [y index=0] {data.csv};
\end{axis}
\end{tikzpicture}
		\caption{Linear}
	\end{subfigure} \hfill	
	\begin{subfigure}{0.32\textwidth}
		\begin{tikzpicture}
\begin{axis}[
    ybar,
ymin=0,
ymax=8,
xmin=29.5,
xmax=36,
    xtick={29,30,31,32,33,34,35,36},
    tick align=outside,
    tick pos=left,
    x grid style={lightgray!92.02614379084967!black},
    y grid style={lightgray!92.02614379084967!black},
    xlabel={rRMSE [\%]},
        width=\figW,
    height=\figH,
]
\addplot +[
    hist={
        bins=13,
        data min=29.5,
        data max=36
    }
] table [y index=0] {data.csv};
\end{axis}
\end{tikzpicture}
		\caption{Local DNN}
	\end{subfigure}	
	\begin{subfigure}{0.32\textwidth}
		\begin{tikzpicture}
\begin{axis}[
    ybar,
ymin=0,
ymax=8,
xmin=33,
xmax=39,
    ymin=0,
    xtick={33,34,35,36,37,38,39},
    tick align=outside,
    tick pos=left,
    x grid style={lightgray!92.02614379084967!black},
    y grid style={lightgray!92.02614379084967!black},
    xlabel={rRMSE [\%]},
        width=\figW,
    height=\figH,
    ylabel={Number of occurrences}
]
\addplot +[
    hist={
        bins=12,
        data min=33,
        data max=39
    }
] table [y index=0] {data.csv};
\end{axis}
\end{tikzpicture}
		\caption{ECMWF}
	\end{subfigure} 	\hfill
	\begin{subfigure}{0.32\textwidth}
		\begin{tikzpicture}
\begin{axis}[
    ybar,
    ybar,
ymin=0,
ymax=8,
xmin=33,
xmax=40,
xtick={33,34,35,36,37,38,39,40},
tick align=outside,
tick pos=left,
x grid style={lightgray!92.02614379084967!black},
y grid style={lightgray!92.02614379084967!black},
xlabel={rRMSE [\%]},
    width=\figW,
height=\figH,
]
\addplot +[
hist={
	bins=14,
	data min=33,
	data max=40
}
] table [y index=0] {data.csv};
\end{axis}
\end{tikzpicture}
		\caption{GBT}
	\end{subfigure} 	\hfill
	\begin{subfigure}{0.32\textwidth}
		\begin{tikzpicture}
\begin{axis}[
    ybar,
    ymin=0,
    ymax=8,
    xmin=40,
    xmax=46,
    xtick={40,41,42,43,44,45,46},
    tick align=outside,
    tick pos=left,
    x grid style={lightgray!92.02614379084967!black},
    y grid style={lightgray!92.02614379084967!black},
    xlabel={rRMSE [\%]},
        width=\figW,
    height=\figH,
]
\addplot +[
    hist={
        bins=12,
        data min=40,
        data max=46
    }
] table [y index=0] {data.csv};
\end{axis}
\end{tikzpicture}
		\caption{Persistent}
	\end{subfigure} 	
\caption{Distribution of the predictive accuracy of the global model across the 25 locations.}
\label{fig:BySiteDist}		
\end{figure*}
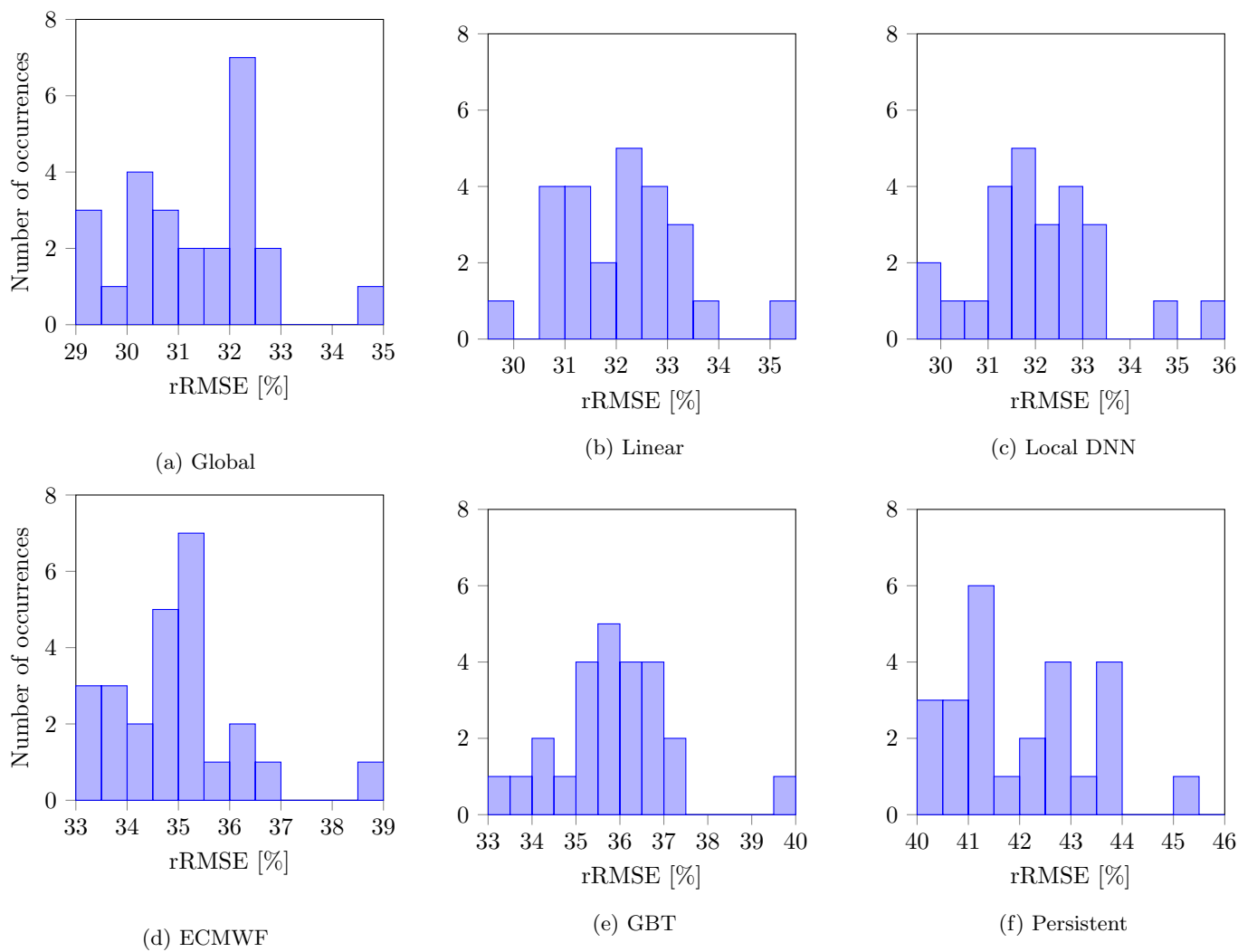

\textred{\subsection{Discussion}
In the previous sections, the performance of the global model has been compared to that of the local models and that of validated models from the literature. Based on the obtained results one can conclude that: (1) the global model is slightly better than the best of the local models; (2) it performs similar to other models from the literature; (3) it provides unbiased forecasts.

While based on these results it cannot be stated that the proposed model is significantly better than all other models, it is important to keep in mind that its main purpose is not to be the best, but to perform equally well as local models so that the operational costs of installing and maintaining a wide sensor network are avoided. In that respect, it can be concluded that the proposed global model is an excellent replacement for the local models: the model is overall slightly better and performs better or equally well across all individual geographical locations and prediction times.}

\section{Conclusion}
\label{sec:conclusion}
In this paper, a general model for short-term forecasting of the global horizontal irradiance has been proposed. The main features of the model are that it replaces ground measurements by satellite-based irradiance values and that, unlike local models previously proposed in the literature, it does not need local measurements in each location where a forecast is needed.

The proposed model was shown to be equal or better than local models typically used in the literature, and in turn, to be an excellent replacement of these local models in order to save the operational costs of installing local sensors and gathering ground data.

In future research, the current work will be expanded with two further investigations. First, the model will be extended to larger regions to analyze whether it generalizes to larger geographical areas than The Netherlands. Second, the model accuracy will be improved by adding other relevant sources of input data, e.g.~weather-based input data like humidity levels or ambient temperature.

\section*{Acknowledgment}
This research has received funding from the European Union’s Horizon 2020 research and innovation program under the Marie Skłodowska-Curie grant agreement No 675318 (INCITE).
\section*{Copyright Information}
\noindent \copyright~2018. This manuscript version is made available under the CC-BY-NC-ND 4.0 license \url{http://creativecommons.org/licenses/by-nc-nd/4.0/}.
\vspace{0.25em}

\noindent \hspace{-0.4em}\doclicenseImage[imagewidth=6em]
\section*{References}

\bibliography{bibtex/bibtex}

\begin{thebibliography}{10}
\expandafter\ifx\csname url\endcsname\relax
  \def\url#1{\texttt{#1}}\fi
\expandafter\ifx\csname urlprefix\endcsname\relax\def\urlprefix{URL }\fi
\expandafter\ifx\csname href\endcsname\relax
  \def\href#1#2{#2} \def\path#1{#1}\fi

\bibitem{Lara-Fanego2012}
V.~Lara-Fanego, J.~A. Ruiz-Arias, D.~Pozo-Vázquez, F.~J. Santos-Alamillos,
  J.~Tovar-Pescador, Evaluation of the {WRF} model solar irradiance forecasts
  in {Andalusia (Southern Spain)}, Solar Energy 86~(8) (2012) 2200--2217.
\newblock \href {http://dx.doi.org/10.1016/j.solener.2011.02.014}
  {\path{doi:10.1016/j.solener.2011.02.014}}.

\bibitem{Voyant2017}
C.~Voyant, G.~Notton, S.~Kalogirou, M.-L. Nivet, C.~Paoli, F.~Motte,
  A.~Fouilloy, Machine learning methods for solar radiation forecasting: A
  review, Renewable Energy 105 (2017) 569--582.
\newblock \href {http://dx.doi.org/10.1016/j.renene.2016.12.095}
  {\path{doi:10.1016/j.renene.2016.12.095}}.

\bibitem{Hammer1999}
A.~Hammer, D.~Heinemann, E.~Lorenz, B.~Lückehe, Short-term forecasting of
  solar radiation: a statistical approach using satellite data, Solar Energy
  67~(1) (1999) 139--150.
\newblock \href {http://dx.doi.org/10.1016/S0038-092X(00)00038-4}
  {\path{doi:10.1016/S0038-092X(00)00038-4}}.

\bibitem{Reikard2009}
G.~Reikard, Predicting solar radiation at high resolutions: A comparison of
  time series forecasts, Solar Energy 83~(3) (2009) 342--349.
\newblock \href {http://dx.doi.org/10.1016/j.solener.2008.08.007}
  {\path{doi:10.1016/j.solener.2008.08.007}}.

\bibitem{Law2014}
E.~W. Law, A.~A. Prasad, M.~Kay, R.~A. Taylor, Direct normal irradiance
  forecasting and its application to concentrated solar thermal output
  forecasting – a review, Solar Energy 108 (2014) 287--307.
\newblock \href {http://dx.doi.org/10.1016/j.solener.2014.07.008}
  {\path{doi:10.1016/j.solener.2014.07.008}}.

\bibitem{Diagne2013}
M.~Diagne, M.~David, P.~Lauret, J.~Boland, N.~Schmutz, Review of solar
  irradiance forecasting methods and a proposition for small-scale insular
  grids, Renewable and Sustainable Energy Reviews 27 (2013) 65--76.
\newblock \href {http://dx.doi.org/10.1016/j.rser.2013.06.042}
  {\path{doi:10.1016/j.rser.2013.06.042}}.

\bibitem{Ahmad2015}
A.~Ahmad, T.~N. Anderson, T.~T. Lie, Hourly global solar irradiation
  forecasting for {New Zealand}, Solar Energy 122 (2015) 1398--1408.
\newblock \href {http://dx.doi.org/10.1016/j.solener.2015.10.055}
  {\path{doi:10.1016/j.solener.2015.10.055}}.

\bibitem{Huang2013}
J.~Huang, M.~Korolkiewicz, M.~Agrawal, J.~Boland, Forecasting solar radiation
  on an hourly time scale using a coupled {AutoRegressive} and dynamical system
  ({CARDS}) model, Solar Energy 87 (2013) 136--149.
\newblock \href {http://dx.doi.org/10.1016/j.solener.2012.10.012}
  {\path{doi:10.1016/j.solener.2012.10.012}}.

\bibitem{Yang2015}
D.~Yang, Z.~Ye, L.~H.~I. Lim, Z.~Dong, Very short term irradiance forecasting
  using the {LASSO}, Solar Energy 114 (2015) 314--326.
\newblock \href {http://dx.doi.org/10.1016/j.solener.2015.01.016}
  {\path{doi:10.1016/j.solener.2015.01.016}}.

\bibitem{Mellit2010}
A.~Mellit, A.~M. Pavan, A 24-h forecast of solar irradiance using artificial
  neural network: Application for performance prediction of a grid-connected
  {PV} plant at {Trieste, Italy}, Solar Energy 84~(5) (2010) 807--821.
\newblock \href {http://dx.doi.org/10.1016/j.solener.2010.02.006}
  {\path{doi:10.1016/j.solener.2010.02.006}}.

\bibitem{Lauret2015}
P.~Lauret, C.~Voyant, T.~Soubdhan, M.~David, P.~Poggi, A benchmarking of
  machine learning techniques for solar radiation forecasting in an insular
  context, Solar Energy 112 (2015) 446--457.
\newblock \href {http://dx.doi.org/10.1016/j.solener.2014.12.014}
  {\path{doi:10.1016/j.solener.2014.12.014}}.

\bibitem{McCandless2015}
T.~C. {McCandless}, S.~E. Haupt, G.~S. Young, A model tree approach to
  forecasting solar irradiance variability, Solar Energy 120 (2015) 514--524.
\newblock \href {http://dx.doi.org/10.1016/j.solener.2015.07.020}
  {\path{doi:10.1016/j.solener.2015.07.020}}.

\bibitem{Lorenz2012}
E.~Lorenz, D.~Heinemann, Prediction of solar irradiance and photovoltaic power,
  in: A.~Sayigh (Ed.), Comprehensive Renewable Energy, Elsevier, 2012, pp.
  239--292.
\newblock \href {http://dx.doi.org/10.1016/B978-0-08-087872-0.00114-1}
  {\path{doi:10.1016/B978-0-08-087872-0.00114-1}}.

\bibitem{Perez2010}
R.~Perez, S.~Kivalov, J.~Schlemmer, K.~Hemker, D.~Renné, T.~E. Hoff,
  Validation of short and medium term operational solar radiation forecasts in
  the {US}, Solar Energy 84~(12) (2010) 2161--2172.
\newblock \href {http://dx.doi.org/10.1016/j.solener.2010.08.014}
  {\path{doi:10.1016/j.solener.2010.08.014}}.

\bibitem{Sfetsos2000}
A.~Sfetsos, A.~H. Coonick, Univariate and multivariate forecasting of hourly
  solar radiation with artificial intelligence techniques, Solar Energy 68~(2)
  (200) 169--178.
\newblock \href {http://dx.doi.org/10.1016/S0038-092X(99)00064-X}
  {\path{doi:10.1016/S0038-092X(99)00064-X}}.

\bibitem{Larson2018}
D.~P. Larson, C.~F.~M. Coimbra, Direct power output forecasts from remote
  sensing image processing, Journal of Solar Energy Engineering 140~(2) (2018)
  021011--021011--8.
\newblock \href {http://dx.doi.org/10.1115/1.4038983}
  {\path{doi:10.1115/1.4038983}}.

\bibitem{Goodfellow2016}
I.~Goodfellow, Y.~Bengio, A.~Courville, Deep Learning, MIT Press, 2016,
  \url{http://www.deeplearningbook.org/}.

\bibitem{Hinton2006}
G.~E. Hinton, S.~Osindero, Y.-W. Teh, A fast learning algorithm for deep belief
  nets, Neural Computation 18~(7) (2006) 1527--1554.
\newblock \href {http://dx.doi.org/10.1162/neco.2006.18.7.1527}
  {\path{doi:10.1162/neco.2006.18.7.1527}}.

\bibitem{Krizhevsky2012}
A.~Krizhevsky, I.~Sutskever, G.~E. Hinton, Imagenet classification with deep
  convolutional neural networks, in: Proceedings of the 25th International
  Conference on Neural Information Processing Systems, NIPS'12, Curran
  Associates Inc., USA, 2012, pp. 1097--1105.
\newblock \href {http://dx.doi.org/10.1145/3065386}
  {\path{doi:10.1145/3065386}}.

\bibitem{Hinton2012}
G.~Hinton, L.~Deng, D.~Yu, G.~E. Dahl, A.~Mohamed, N.~Jaitly, A.~Senior,
  V.~Vanhoucke, P.~Nguyen, T.~N. Sainath, B.~Kingsbury, Deep neural networks
  for acoustic modeling in speech recognition: The shared views of four
  research groups, Signal Processing Magazine 29~(6) (2012) 82--97.
\newblock \href {http://dx.doi.org/10.1109/MSP.2012.2205597}
  {\path{doi:10.1109/MSP.2012.2205597}}.

\bibitem{Bahdanau2014}
D.~Bahdanau, K.~Cho, Y.~Bengio, Neural machine translation by jointly learning
  to align and translate, arXiv eprint (2014).
\newblock \href {http://arxiv.org/abs/1409.0473} {\path{arXiv:1409.0473}}.

\bibitem{Wang2016}
H.~Wang, G.~Wang, G.~Li, J.~Peng, Y.~Liu, Deep belief network based
  deterministic and probabilistic wind speed forecasting approach, Applied
  Energy 182 (2016) 80--93.
\newblock \href {http://dx.doi.org/10.1016/j.apenergy.2016.08.108}
  {\path{doi:10.1016/j.apenergy.2016.08.108}}.

\bibitem{Feng2017}
C.~Feng, M.~Cui, B.-M. Hodge, J.~Zhang, A data-driven multi-model methodology
  with deep feature selection for short-term wind forecasting, Applied Energy
  190 (2017) 1245--1257.
\newblock \href {http://dx.doi.org/10.1016/j.apenergy.2017.01.043}
  {\path{doi:10.1016/j.apenergy.2017.01.043}}.

\bibitem{Suryanarayana2018}
G.~Suryanarayana, J.~Lago, D.~Geysen, P.~Aleksiejuk, C.~Johansson, Thermal load
  forecasting in district heating networks using deep learning and advanced
  feature selection methods, Energy 157 (2018) 141--149.
\newblock \href {http://dx.doi.org/10.1016/j.energy.2018.05.111}
  {\path{doi:10.1016/j.energy.2018.05.111}}.

\bibitem{Coelho2017}
I.~Coelho, V.~Coelho, E.~Luz, L.~Ochi, F.~Guimarães, E.~Rios, A {GPU} deep
  learning metaheuristic based model for time series forecasting, Applied
  Energy 201 (2017) 412--418.
\newblock \href {http://dx.doi.org/10.1016/j.apenergy.2017.01.003}
  {\path{doi:10.1016/j.apenergy.2017.01.003}}.

\bibitem{Fan2017}
C.~Fan, F.~Xiao, Y.~Zhao, A short-term building cooling load prediction method
  using deep learning algorithms, Applied Energy 195 (2017) 222--233.
\newblock \href {http://dx.doi.org/10.1016/j.apenergy.2017.03.064}
  {\path{doi:10.1016/j.apenergy.2017.03.064}}.

\bibitem{Lago2018}
J.~Lago, F.~{De Ridder}, P.~Vrancx, B.~{De Schutter}, Forecasting day-ahead
  electricity prices in {Europe}: The importance of considering market
  integration, Applied Energy 211 (2018) 890--903.
\newblock \href {http://dx.doi.org/10.1016/j.apenergy.2017.11.098}
  {\path{doi:10.1016/j.apenergy.2017.11.098}}.

\bibitem{Lago2018a}
J.~Lago, F.~{De Ridder}, B.~{De Schutter}, Forecasting spot electricity prices:
  deep learning approaches and empirical comparison of traditional algorithms,
  Applied Energy 221 (2018) 386--405.
\newblock \href {http://dx.doi.org/10.1016/j.apenergy.2018.02.069}
  {\path{doi:10.1016/j.apenergy.2018.02.069}}.

\bibitem{Weron2014}
R.~Weron, {Electricity price forecasting: A review of the state-of-the-art with
  a look into the future}, International Journal of Forecasting 30~(4) (2014)
  1030--1081.
\newblock \href {http://dx.doi.org/10.1016/j.ijforecast.2014.08.008}
  {\path{doi:10.1016/j.ijforecast.2014.08.008}}.

\bibitem{Ruder2016}
S.~Ruder, An overview of gradient descent optimization algorithms, arXiv eprint
  (2016).
\newblock \href {http://arxiv.org/abs/1609.04747} {\path{arXiv:1609.04747}}.

\bibitem{Bergstra2011}
J.~Bergstra, R.~Bardenet, Y.~Bengio, B.~K{\'{e}}gl, Algorithms for
  hyper-parameter optimization, in: Advances in Neural Information Processing
  Systems, 2011, pp. 2546--2554.

\bibitem{Hutter2011}
F.~Hutter, H.~H. Hoos, K.~Leyton-Brown, Sequential model-based optimization for
  general algorithm configuration, in: International Conference on Learning and
  Intelligent Optimization, Springer, 2011, pp. 507--523.
\newblock \href {http://dx.doi.org/10.1007/978-3-642-25566-3_40}
  {\path{doi:10.1007/978-3-642-25566-3_40}}.

\bibitem{Marquez2012}
R.~Marquez, C.~F.~M. Coimbra, Proposed metric for evaluation of solar
  forecasting models, Journal of Solar Energy Engineering 135~(1) (2012)
  011016--011016--9.
\newblock \href {http://dx.doi.org/10.1115/1.4007496}
  {\path{doi:10.1115/1.4007496}}.

\bibitem{Ineichen2002}
P.~Ineichen, R.~Perez, A new airmass independent formulation for the {Linke}
  turbidity coefficient, Solar Energy 73~(3) (2002) 151--157.
\newblock \href {http://dx.doi.org/10.1016/S0038-092X(02)00045-2}
  {\path{doi:10.1016/S0038-092X(02)00045-2}}.

\bibitem{Greuell2013}
W.~Greuell, J.~Meirink, P.~Wang, Retrieval and validation of global, direct,
  and diffuse irradiance derived from {SEVIRI} satellite observations, Journal
  of Geophysical Research: Atmospheres 118~(5) (2013) 2340--2361.
\newblock \href {http://dx.doi.org/10.1002/jgrd.50194}
  {\path{doi:10.1002/jgrd.50194}}.

\bibitem{Deneke2008}
H.~Deneke, A.~Feijt, R.~Roebeling, Estimating surface solar irradiance from
  {METEOSAT} {SEVIRI}-derived cloud properties, Remote Sensing of Environment
  112~(6) (2008) 3131--3141.
\newblock \href {http://dx.doi.org/10.1016/j.rse.2008.03.012}
  {\path{doi:10.1016/j.rse.2008.03.012}}.

\bibitem{Srivastava2014}
N.~Srivastava, G.~Hinton, A.~Krizhevsky, I.~Sutskever, R.~Salakhutdinov,
  Dropout: A simple way to prevent neural networks from overfitting, Journal of
  Machine Learning Research 15 (2014) 1929--1958.

\bibitem{Nair2010}
V.~Nair, G.~E. Hinton, Rectified linear units improve restricted boltzmann
  machines, in: Proceedings of the 27th international Conference on Machine
  Learning (ICML), 2010, pp. 807--814.

\bibitem{Kingma2014}
D.~P. Kingma, J.~Ba, Adam: A method for stochastic optimization, arXiv eprint
  (2014).
\newblock \href {http://arxiv.org/abs/1412.6980} {\path{arXiv:1412.6980}}.

\bibitem{Yao2007}
Y.~Yao, L.~Rosasco, A.~Caponnetto, On early stopping in gradient descent
  learning, Constructive Approximation 26~(2) (2007) 289--315.
\newblock \href {http://dx.doi.org/10.1007/s00365-006-0663-2}
  {\path{doi:10.1007/s00365-006-0663-2}}.

\bibitem{ecmwf}
{European Centre for Medium-Range Weather Forecasts (ECMWF)} website,
  \url{https://www.ecmwf.int/}.

\bibitem{knmi}
\href{http://knmi.nl/}{{Royal Netherlands Meteorological Institute (KNMI)}}.
\newline\urlprefix\url{http://knmi.nl/}

\bibitem{Andrews2014}
R.~W. Andrews, J.~S. Stein, C.~Hansen, D.~Riley, Introduction to the open
  source {PV LIB} for python photovoltaic system modelling package, in:
  Photovoltaic Specialist Conference (PVSC), 2014 IEEE 40th, IEEE, 2014, pp.
  170--174.
\newblock \href {http://dx.doi.org/10.1109/pvsc.2014.6925501}
  {\path{doi:10.1109/pvsc.2014.6925501}}.

\bibitem{Chen2016}
T.~Chen, C.~Guestrin, Xgboost: A scalable tree boosting system, in: Proceedings
  of the 22nd {ACM SIGKDD} International Conference on Knowledge Discovery and
  Data Mining, 2016, pp. 785--794.

\bibitem{Hastie2001}
T.~Hastie, R.~Tibshirani, J.~Friedman, The Elements of Statistical Learning,
  Springer Series in Statistics, Springer New York Inc., New York, NY, USA,
  2001.
\newblock \href {http://dx.doi.org/10.1007/978-0-387-21606-5}
  {\path{doi:10.1007/978-0-387-21606-5}}.

\end{thebibliography}

\end{document}